\documentclass[pdflatex,sn-mathphys-num]{sn-jnl}

\usepackage{graphicx}%
\graphicspath{{figures/}} 
\usepackage{multirow}%
\usepackage{amsmath,amssymb,amsfonts}%
\usepackage{amsthm}%
\usepackage{mathrsfs}%
\usepackage[title]{appendix}%
\usepackage{xcolor}%
\usepackage{textcomp}%
\usepackage{manyfoot}%
\usepackage{booktabs}%
\usepackage{algorithm}%
\usepackage{algorithmicx}%
\usepackage{algpseudocode}%
\usepackage{listings}%
\usepackage{subcaption}%
\usepackage{wrapfig}%
\usepackage{pgfplots}%
\pgfplotsset{compat=1.14}%
\usepackage{makecell}%

\theoremstyle{thmstyleone}%
\theoremstyle{thmstyletwo}%

\theoremstyle{thmstylethree}%

\begin{document}

\title{DROM: A Language-Guided Diffusion Framework for Multi-Skill Robotic Manipulation}


\author*[1]{\fnm{Vincenzo} \sur{Pomponi}}\email{vincenzo.pomponi@supsi.ch}
\author[1]{\fnm{Rocco} \sur{Felici}}\email{rocco.felici@supsi.ch}
\author[2]{\fnm{Paolo} \sur{Franceschi}}\email{paolo.franceschi@supsi.ch}
\author[1]{\fnm{Stefano} \sur{Baraldo}}\email{stefano.baraldo@supsi.ch}
\author[1]{\fnm{Oliver} \sur{Avram}}\email{oliver.avram@supsi.ch}
\author[2,3]{\fnm{Loris} \sur{Roveda}}\email{loris.roveda@supsi.ch} \email{ loris.roveda@polimi.it}
\author[4]{\fnm{Luca Maria} \sur{Gambardella}}\email{luca.gambardella@usi.ch}
\author[1]{\fnm{Anna} \sur{Valente}}\email{anna.valente@supsi.ch}

\affil[1]{\orgdiv{Institute of Systems and Technologies for Sustainable Production (ISTePS), Department of Innovative Technologies}, \orgname{University of Applied Science and Arts of Southern Switzerland (SUPSI)}, \orgaddress{\street{Via la Santa 1}, \city{Lugano}, \postcode{CH-6900}, \state{Ticino}, \country{Switzerland}}}

\affil[2]{\orgdiv{Istituto Dalle Molle di studi sull’intelligenza artificiale (IDSIA), Department of Innovative Technologies}, \orgname{University of Applied Science and Arts of Southern Switzerland (SUPSI)}, \orgaddress{\street{Via la Santa 1}, \city{Lugano}, \postcode{CH-6900}, \state{Ticino}, \country{Switzerland}}}

\affil[3]{\orgdiv{Mechanical Department}, \orgname{Politecnico di Milano (PoliMi)}, \orgaddress{\street{Via Giuseppe Candiani}, \city{Milan}, \postcode{20155}, \country{Italy}}}

\affil[4]{\orgdiv{Istituto Dalle Molle di studi sull’intelligenza artificiale (IDSIA), Faculty of Informatics}, \orgname{Università della Svizzera Italiana (USI)}, \orgaddress{\street{Via Buffi 13}, \city{Lugano}, \postcode{CH-6900}, \state{Ticino}, \country{Switzerland}}}


\abstract{
Learning robust manipulation policies for diverse, long-horizon tasks from limited demonstrations remains a fundamental challenge in robotics.
We present \textit{DROM}, a language-guided diffusion framework that enables robots to learn, represent, and compose multiple manipulation skills within a single generative policy.
DROM leverages Dynamic Movement Primitives (DMPs) to augment a small set of expert demonstrations into expressive multi-skill datasets, substantially reducing data collection while improving spatial generalization beyond the demonstrated workspace.
Building upon Motion Planning Diffusion (MPD), we extend the diffusion architecture to support language-conditioned multi-skill trajectory generation through cross-attention, allowing a single model to generate skill-consistent motions for a diverse set of manipulation primitives, including orientation-sensitive behaviors that are difficult to design using conventional motion planning or hard-coded controllers.
For long-horizon manipulation, a large language model decomposes high-level operator requests into executable sequences of skills, enabling natural language interaction and autonomous task execution.
We validate DROM on a Franka Emika Panda robot, a FANUC CRX25ia robot, and in MuJoCo simulation across a wide range of manipulation tasks.
Experimental results demonstrate that DROM outperforms Motion Planning Diffusion and Behavior Cloning baselines, achieves robust multi-skill generalization, and composes learned skills to reliably execute long-horizon manipulation tasks from natural language instructions using only a limited number of human demonstrations.
Datasets, simulation environments, and more at https://github.com/automation-robotics-machines/drom.
}




\maketitle

\begin{figure}[ht]
    \centering
    \includegraphics[width=1.0\linewidth]{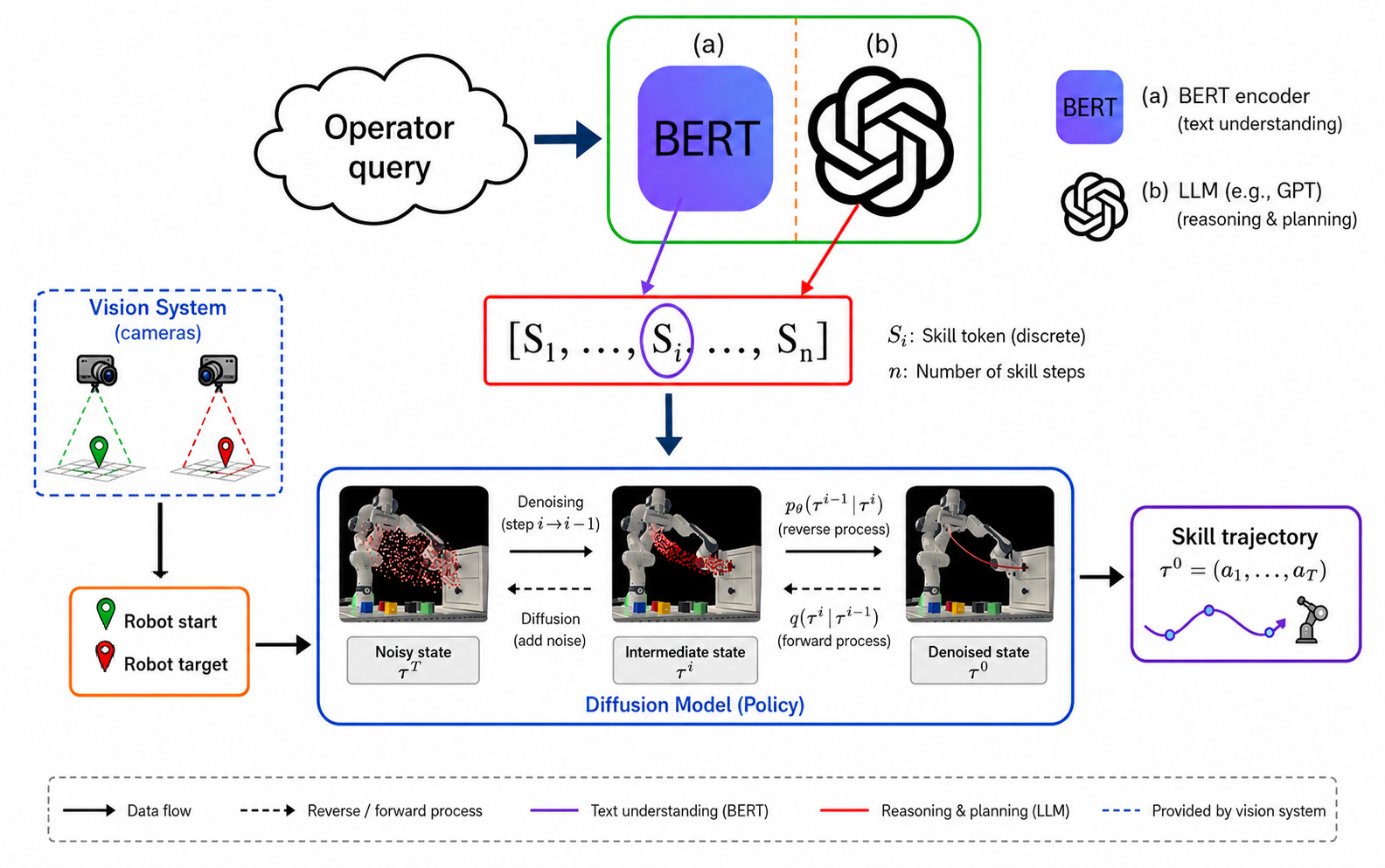}
    \caption{\textbf{Illustration of the two operational modalities of DROM:} (a) language-guided single-skill execution, where an operator query is embedded and mapped to a manipulation primitive to generate a skill-consistent trajectory; and (b) language-guided long-horizon planning, where a language model decomposes a high-level objective into an ordered sequence of skills, each executed via diffusion-based trajectory generation.}
    \label{fig:model_schema}
\end{figure}

\section{Introduction}
\label{sec:introduction}

Developing robotic control policies that operate reliably in unstructured environments remains a core challenge in manipulation learning~\cite{ROVEDA202368}.
Most conventional imitation learning approaches require large, task-specific datasets and are typically limited to individual manipulation skills, limiting scalability and reuse across broader operational workflows~\cite{felici2023imitation}.
Furthermore, while these methods can replicate demonstrated behaviors, they frequently struggle to generalize when encountering workspace and object configurations that deviate significantly from the training distribution~\cite{kawaharazuka2024real}.

Foundation Models have emerged as a promising paradigm to address these limitations by leveraging strong generative and multimodal reasoning capabilities~\cite{osa2018algorithmic,team2024octo}.
In particular, Vision-Language-Action (VLA) models~\cite{kawaharazuka2025vision} integrate perception, language understanding, and action generation to execute long-horizon tasks directly from high-level instructions~\cite{brohan2023rt,yu2025survey,kim2024openvla}.
However, these end-to-end architectures typically require massive internet-scale datasets and rely on computationally heavy inference pipelines, introducing significant latency that can hinder real-time deployment and reactive control on physical hardware~\cite{bommasani2021opportunities,firoozi2025foundation}.

To bypass these operational bottlenecks, recent advances have focused on continuous generative models for direct data trajectory synthesis.
Diffusion Models~\cite{ho2020denoising} and Flow Matching frameworks~\cite{rouxel2024flow} have proven highly effective at capturing complex, multimodal motion distributions~\cite{chi2023diffusion}.
While Flow Matching offers faster inference capabilities~\cite{nguyen2025flowmp}, iterative diffusion models provide stronger regularization properties through their multi-step denoising process, which yields superior trajectory-level consistency and robustness under unseen conditions~\cite{wolf2025diffusion,wang2022diffusion}.
Nevertheless, existing diffusion-based manipulation policies are commonly restricted to single skills, require dedicated data collection pipelines for every unique subtask, and remain inherently sample-inefficient when trained from scratch on raw human demonstrations.

Consequently, a critical gap remains: creating a scalability bridge that satisfies the dense data requirements of expressive diffusion models without demanding exhaustive human demonstration campaigns, while simultaneously executing language-guided long-horizon tasks without introducing language-model latency into the low-level control loop.

To address these challenges, we introduce \textbf{DROM} (\textbf{D}iffusion for \textbf{RO}botic \textbf{M}anipulation), a language-guided, multi-skill diffusion framework for robotic manipulation.
DROM combines a data-efficient training pipeline with hierarchical task reasoning.
First, it leverages Dynamic Movement Primitives (DMPs) to transform a small set of human demonstrations into a large synthetic multi-skill dataset, enabling the diffusion model to learn expressive trajectory distributions and generalize beyond the demonstrated workspace.
Second, high-level natural language instructions are interpreted before execution to identify either a single manipulation primitive or a sequence of skills, while the diffusion model generates the corresponding trajectories from the current environment state.
This modular design decouples task reasoning from motion generation and enables efficient learning from synthetic demonstrations, providing a scalable solution for language-guided, multi-skill robotic manipulation.

\section{Related Works}
\label{sec:related_works}
\subsection{DMPs and Synthetic Trajectory Generation}
Learning from Demonstration (LfD) enables non-expert users to teach robots through example executions~\cite{Calinon2018, ROVEDA202368}.
Among LfD approaches, Dynamic Movement Primitives (DMPs) represent demonstrations as nonlinear dynamical systems, supporting trajectory reproduction, adaptation to novel goals, and guaranteed convergence~\cite{dmps01, dmps02, dmps03, 5152385}.
Their flexibility has been further extended through obstacle-aware modulation~\cite{5152423} and Coupling Movement Primitives~\cite{6748918}.
While DMPs have been extensively employed in imitation learning, their use as a mechanism for generating datasets that encode complex, object-specific manipulation behaviors that are difficult to hard-code has remained largely unexplored.
Starting from a small set of demonstrations, we generate a structured and diverse trajectory dataset for robotic manipulation, similarly to~\cite{pomponi2026dynamimicgen}.
The diffusion model subsequently learns the underlying manipulation behavior from these augmented trajectories and generalizes it beyond the training distribution, enabling robust trajectory generation over an extended operational workspace.

\subsection{Diffusion Models for Trajectory Planning}
Learning multi-task robotic manipulation in complex environments is crucial for scalable robotic systems, and recent advances in deep generative models have shown strong capabilities in capturing complex motion distributions~\cite{yan2024dnact}.
Diffusion Models, in particular, are well suited for trajectory generation, as they preserve the global structure of motions while enabling diverse behaviors and stable training~\cite{ho2020denoising}.
Several works have explored the application of diffusion models to robotic tasks.
Pearce et al.~\cite{pearce2023imitating}, investigate different architectures to adapt diffusion models for human behavior imitation, closely matching demonstrated trajectories and motivating our choice of a temporal U-Net architecture.
Similarly, Janner et al.~\cite{janner2022planning} apply diffusion models for planning and trajectory inference across various environments, further confirming their effectiveness for robotic motion generation.
Movement Primitive Diffusion~\cite{scheikl2024movement} demonstrated the potential synergy between DMPs within the diffusion framework by predicting DMP parameters that are decoded into smooth trajectories.
In contrast, our approach leverages DMPs and diffusion models separately: using DMPs for structured trajectory augmentation and diffusion models for direct trajectory synthesis.
Other works have emphasized multi-task and multi-modal conditioning.
For instance, Chen et al.~\cite{chen2024diffusion} demonstrate information sharing across tasks during the diffusion process using task-conditional probabilities in reverse denoising, which inspired our multi-task-conditioned diffusion design.
Hierarchical Diffusion Policy (HDP)~\cite{ma2024hierarchical} separates multi-task manipulation into a high-level planner predicting next-best end-effector poses and a low-level diffusion policy generating corresponding trajectories, motivating our integration of a high-level language model to generate sequences of skills for long-horizon goals.

\subsection{Skill Composition for Long-Horizon Tasks}
Skill composition is commonly integrated within motion planning frameworks~\cite{kaelbling2011hierarchical, urbaniak2021combining} to enable robotic systems to operate effectively in multi-task and unstructured environments.
Such approaches typically rely on search-based or hierarchical planning strategies to determine sequences of actions that achieve a desired goal~\cite{shaoul2024multi}.
However, these approaches require extensive pre-programming and lack the ability to generalize to new unscripted tasks~\cite{ahn2022can}.
More recently, the literature has emphasized the importance of structuring task plans directly from natural language instructions, allowing users to interact with robotic systems in a more intuitive and flexible manner~\cite{lynch2023interactive, yang2024guidinglonghorizontaskmotion, liu2025learningcompositionalbehaviorsdemonstration}.

Motivated by this perspective, we propose a framework that leverages language models to translate high-level user instructions into an ordered sequence of executable motions, enabling the achievement of a long-horizon objective. 

Inspired by recent developments in Learning from Demonstration (LfD) and skill composition, we extend the MPD formulation~\cite{carvalho2025motion} by transforming single-skill diffusion planning into a unified multi-skill framework.
Building on prior work, our approach incorporates a language encoder to embed operator queries, enabling the recall of the same skill from different prompts, and a language model that combines hierarchical task decomposition with generative trajectory modeling to solve diverse, long-horizon robotic manipulation tasks.

\bigskip

In particular, we introduce DROM with the following key contributions:
\begin{enumerate}
    \item \textbf{DMP-Augmented Data Pipeline:} We design a structured data generation pipeline that expands a limited set of human demonstrations into diverse, skill-consistent training trajectories for diffusion-based manipulation learning.
    By utilizing Dynamic Movement Primitives (DMPs) to mathematically bootstrap expert demonstrations for each skill, the framework expands sparse human inputs into dense, workspace-wide datasets capable of capturing contact-rich behaviors for challenging objects.
    \item \textbf{Unified Multi-Skill Diffusion Model:} We extend the single-task architecture of MPD~\cite{carvalho2025motion} into a single, unified framework capable of handling 17 distinct manipulation primitives simultaneously.
    By embedding a cross-attention layer into a shared temporal U-Net backend, the model learns task-specific behaviors while generalizing beyond the training distribution.
    \item \textbf{Language-Conditioned Trajectory Generation:} We introduce a dual language-conditioning framework that combines a lightweight language encoder with a high-level language model. Confidence thresholds computed through clustered conformal prediction determine whether a user query requires a single manipulation primitive or to be decomposed into an ordered sequence of skills for long-horizon task execution.
    \item \textbf{Cross-Platform Validation:} We perform extensive empirical evaluations in MuJoCo simulation and on two distinct physical manipulators (a Franka Emika Panda and a FANUC CRX25ia). DROM consistently outperforms baseline configurations of MPD~\cite{carvalho2025motion} and Behavior Cloning~\cite{torabi2018behavioral}, demonstrating robust spatial extrapolation and reliable skill consistency.
\end{enumerate}

\section{Method}
\label{sec:method}

This section presents the DROM framework (Fig.~\ref{fig:model_schema}), which combines language understanding and diffusion-based trajectory generation to execute both single-skill and long-horizon manipulation tasks.

First, a small number of expert demonstrations is leveraged to generate a large synthetic dataset through the proposed DMP-based augmentation pipeline.
The resulting trajectories are used to train a language-conditioned diffusion model, while a language encoder and a large language model (LLM) are prepared to interpret operator instructions.

During inference, clustered conformal prediction \cite{ding2023classconditionalconformalpredictionclasses} determines whether the user's instruction corresponds to a known manipulation primitive or a high-level task (Sec.~\ref{subsubsec:conf_pred}).
In the former case, the language encoder directly identifies the requested skill, which conditions the diffusion model for trajectory generation.
Otherwise, the instruction is forwarded to the LLM, which decomposes it into an ordered sequence of learned manipulation skills.
For each selected skill, the diffusion model generates executable Cartesian trajectories and gripper commands conditioned on the current robot and object states obtained through proprioception and vision.

\subsection{Diffusion-based Planning}
\label{subsec:diff_planning}
\begin{figure}[ht]
    \centering
    \includegraphics[width=1.0\linewidth]{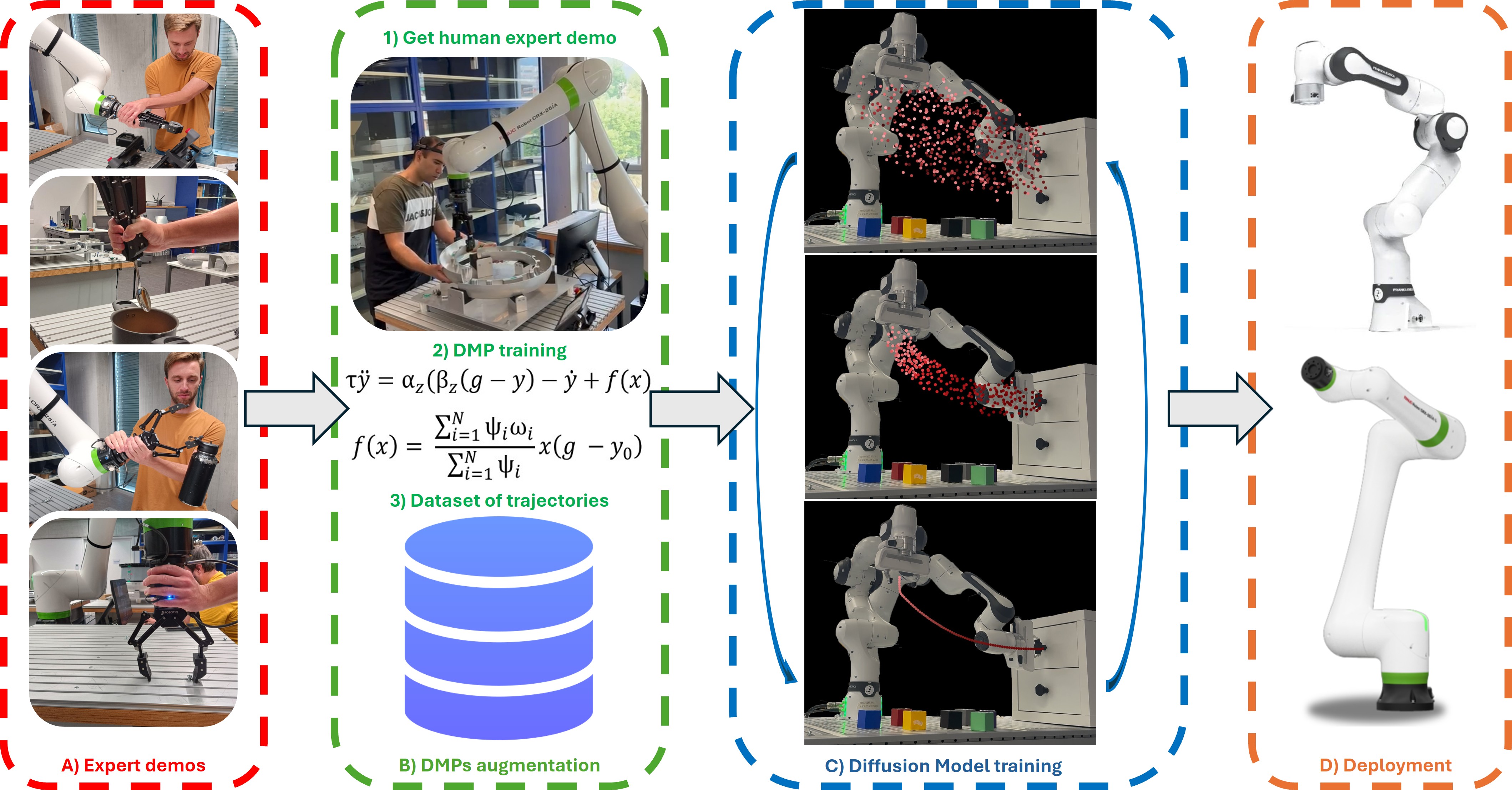}
    \caption{Overview of the diffusion-based training pipeline of DROM: (A) collection of a single expert demonstration per skill; (B) synthetic dataset expansion using Dynamic Movement Primitives (DMPs); (C) training of the skill-conditioned diffusion model on the augmented multi-skill dataset; and (D) deployment in simulation and on real robotic platforms.}
    \label{fig:inference}
\end{figure}
This section presents the diffusion-based planning pipeline of DROM, illustrated in Fig.~\ref{fig:inference}.
The pipeline consists of four stages: (i) expert demonstration acquisition through kinesthetic teaching or teleoperation, (ii) DMP-based synthetic data augmentation, (iii) training of the skill-conditioned diffusion model, and (iv) deployment on both simulated and real robotic platforms.

\subsubsection{Data Collection}
\label{subsubsec:data_collection}
The first stage of the diffusion model training consists of acquiring expert demonstrations for each manipulation skill (Fig.~\ref{fig:inference}A).
Demonstrations are collected through kinesthetic teaching on the real robotic platforms (FANUC CRX25ia and Franka Emika Panda) and through a SpaceMouse interface in simulation (MuJoCo).

For each skill $k \in \{1,\ldots,K\}$, the expert provides $L_k \geq 1$ demonstrations represented as $\mathbf{d}^l \in \mathbb{R}^{H \times J}$, where $l$ is a global demonstration index, $H$ is the trajectory horizon, $J$ is the waypoint dimensionality, and $K$ is the total number of skills.
Following~\cite{carvalho2025motion}, all trajectories are represented with a fixed horizon of $H=64$.
Each waypoint comprises the Cartesian end-effector position $(P_x,P_y,P_z)$, orientation encoded as a quaternion $(Q_x,Q_y,Q_z,Q_w)$, and a binary gripper command $G\in\{-1,+1\}$ indicating the gripper state.

The complete demonstration set is defined as
\[
\mathcal{D}_{\text{demo}} = \{\mathbf{d}^1,\ldots,\mathbf{d}^l,\ldots,\mathbf{d}^L\},
\]
where $L=\sum_k L_k \geq K$ denotes the total number of expert demonstrations.
Multiple trajectory demonstrations may be provided by the human operator to encode the required manipulation behavior when a single demonstration is insufficient to capture the variability of the skill.
In particular, additional demonstrations may be provided for tasks involving challenging object-specific interaction patterns, where the approach strategy varies significantly with the object pose.
This is especially relevant for tasks such as grasping bottles or cups, where the end-effector orientation and approach direction play a critical role in successful execution.


\subsubsection{Data Augmentation}
\label{subsubsec:dmps_augmentation}

Effective training of diffusion-based imitation learning models requires a sufficiently diverse dataset capable of capturing the variability associated with different manipulation skills.
Since the initial dataset $D_{\text{demo}}$ consists of a limited number of expert demonstrations, we employ Dynamic Movement Primitives (DMPs) to synthetically augment the available data, as illustrated in Fig.~\ref{fig:inference}B.
Further details on the DMP formulation and its mathematical derivation are provided in Appendix~\ref{app:DMPs}.

As described in Section~\ref{subsubsec:data_collection}, skills requiring more diverse manipulation strategies are represented by multiple expert demonstrations.
Each demonstration $\mathbf{d}^l$ is encoded by a corresponding Dynamic Movement Primitive, $\text{DMP}^l$, which synthesizes trajectory variations by modifying the start and goal states while maintaining the underlying manipulation behavior.

The resulting training dataset contains $M \times K$ trajectories:
\[
\mathcal{D}_{\text{train}} = \{\mathcal{D}_1, \ldots, \mathcal{D}_k, \ldots, \mathcal{D}_K\},
\]
where $\mathcal{D}_k \in \mathbb{R}^{H \times J \times M}$ represents the synthetic dataset of trajectories generated for the skill $k$.
In all experiments, each skill is represented by a fixed dataset of $M=300$ synthetic trajectories, which we found sufficient to provide the diversity required for robust diffusion model training.
When $L_k > 1$ expert demonstrations are available for the same skill $k$, the trajectory budget is equally allocated among the corresponding DMPs so that the overall number of generated trajectories per skill remains equal to $M$.

\subsubsection{Skill-Conditioned Diffusion Model}
\label{subsubsec:diffusion_model}

Diffusion Models~\cite{ho2020denoising} (Fig.~\ref{fig:inference}C) define a forward stochastic process that gradually corrupts trajectories sampled from the data distribution, $\tau_0 \sim q(\tau_0)$, into isotropic Gaussian noise.

This process is modeled as a Markov chain:
\begin{equation}
q(\tau_t \mid \tau_{t-1}, t) = \mathcal{N}\big(\tau_t;\, \sqrt{1-\beta_t}\,\tau_{t-1},\, \beta_t \mathbf{I}\big), \quad t = 1, \dots, N,
\end{equation}
where $N$ denotes the total number of diffusion steps and $\beta_t$ controls the noise schedule at each step.

Under the assumption that trajectories lie in a Euclidean space, the marginal distribution at an arbitrary timestep admits a closed-form expression:
\begin{equation}
q(\tau_t \mid \tau_0, t) = \mathcal{N}\big(\tau_t;\, \sqrt{\bar{\alpha}_t}\,\tau_0,\,(1 - \bar{\alpha}_t)\mathbf{I}\big),
\end{equation}
where
\begin{equation}
\bar{\alpha}_t = \prod_{s=1}^{t} (1 - \beta_s).
\end{equation}

The reverse process aims to recover data trajectories from Gaussian noise via a learned denoising model:
\begin{equation}
p_\theta(\tau_{t-1} \mid \tau_t, t) = \mathcal{N}\big(\tau_{t-1};\, \mu_\theta(\tau_t, t),\, \Sigma_t\big),
\end{equation}
where the mean $\mu_\theta$ is learned, while the covariance is fixed according to the forward process:
\begin{equation}
\Sigma_t = \tilde{\beta}_t \mathbf{I}, \quad \tilde{\beta}_t = \beta_t \frac{1 - \bar{\alpha}_{t-1}}{1 - \bar{\alpha}_t}.
\end{equation}

Following Ho et al.~\cite{ho2020denoising}, we instead parameterize the model to predict the noise term $\epsilon_\theta(\tau_t, t)$, leading to the simplified training objective:
\begin{equation}
\mathcal{L}(\theta) = \mathbb{E}_{t, \epsilon, \tau_0} \left[ \lVert \epsilon - \epsilon_\theta(\tau_t, t) \rVert_2^2 \right],
\end{equation}
where
\begin{equation}
t \sim \mathcal{U}(1, N), \quad \epsilon \sim \mathcal{N}(0, \mathbf{I}), \quad \tau_t = \sqrt{\bar{\alpha}_t}\tau_0 + \sqrt{1 - \bar{\alpha}_t}\epsilon.
\end{equation}

To enable skill-conditioned trajectory generation, we condition the diffusion process on a learned skill embedding $f_\theta(p)$ encoding the primitive $p$:
\begin{equation}
\epsilon_\theta(\tau_t, t, f_\theta(p)).
\end{equation}

This conditioning allows a single diffusion model to generate trajectories for multiple skills by injecting task-specific information at every reverse diffusion step.

For trajectory modeling, we adopt the temporal U-Net architecture proposed by Carvalho et al.~\cite{carvalho2025motion}, which has shown strong performance in motion generation tasks. We further extend this architecture by integrating a cross-attention mechanism to incorporate skill conditioning within the diffusion process.

\subsection{Language-based Planning}
\label{subsec:lang_planning}

This section presents the language-based planning part of DROM, which interprets natural language instructions and determines whether they correspond to a single manipulation primitive or a long-horizon task.
To this end, clustered conformal prediction is used to distinguish between the two cases.
Single-skill requests are directly mapped to the corresponding primitive, whereas high-level instructions are decomposed by a language model into an ordered sequence of executable skills.

\subsubsection{Conformal Prediction for Primitive Classification} 
\label{subsubsec:conf_pred}
DROM has been designed to provide a set of $K$ skills $\mathcal{P}=\{p_k\}_{k=1}^K$.
Yet a user, interacting with the system, might not only request simple skills among the available ones, but also long horizon objectives that would require a sequence of skills or even goals outside the scope of the robot's capabilities. 
To discriminate among these cases and define to what the utterance of the user refers to, a clustered conformal procedure~\cite{ding2023classconditionalconformalpredictionclasses} has been used.
A set of thresholds $\mathcal{L}=\{\lambda_k\}_{k=1}^K$ is defined, where each $\lambda_k$ is associated with one skill $p_k$.

During an offline phase, a set of $U_k$ examples of utterances for each skill $p_k$ is collected: $\mathcal{U}^k=\{u_i^k\}_{i=1}^{U_k}$.
Given a pretrained language embedding model $f_\theta$ with parameters $\theta$, the set $\mathcal{E}^k=\{e_i^ k\}_{i=1}^{U_k}=\{f_\theta(u_i^ k)\}_{i=1}^{U_k}$ of latent embeddings $e_i^k\in\mathbb{R}^d$ of each collected utterance $u_i^k$ is created. 



Then conformal scores are defined as the maximum cosine similarity observed in pairwise comparisons between examples of the same skill:
\begin{equation}\label{eq:1}
\text{score}(e_i^k, \mathcal{E}^k) = \max_{e_j^k \in \mathcal{E}^k\setminus\{e^k_i\} } S_\text{cos}(e_i^k, e_j^k), \quad \forall i
\end{equation}
with
$S_{\text{cos}}(e_1,e_2) = \widehat{e_1}^\top\widehat{e_2}$, where generic embeddings $e_1,e_2$ are L2-normalized, i.e.,  $\widehat{e_1}=\frac{e_1}{\lVert e_1 \rVert_2}$, $\widehat{e_2}=\frac{e_2}{\lVert e_2 \rVert_2}$.
Skills are then clustered into $h$ groups $\{c_1,\dots,c_h\}$ according to the similarity of their score distributions.
Then at cluster level, for a level $\alpha$, the quantile $\hat{q}$ of the scores is estimated.
Thresholds for skills are set to the quantiles of the groups they were assigned to:
\begin{equation}\label{eq:2}\lambda_k = \hat{q}_{c_l}\quad \text{s.t.} \quad p_k \in c_l, \quad \forall p_k \in \mathcal{P}.
\end{equation}

At run time, given the embedding $e_q$ of the query $q$ of the operator, DROM responds differently to three possible scenarios:
\begin{itemize}
    \item[]($i$) unscripted or out-of-distribution tasks: no given skill is similar to the operator's query;
    \item[]($ii$) classification uncertainty: two or more skills are likely similar to the operator's query;
    \item[]($iii$) among all, only one skill is similar to the instruction of the operator.
\end{itemize}
In the first scenario ($i$), if $\text{score}(e_q, \mathcal{E}^k)$ is greater than the corresponding threshold $\lambda_k$ the query will most likely not refer to that skill $p_k$; if this is observed for all skills, the utterance will be considered an unscripted task or an out-of-distribution task to be subjected to the LLM generation (Sec.~\ref{subsubsec:llm}).
In the second scenario ($ii$), when the scores of multiple skills fall below their corresponding thresholds, all such skills are considered plausible candidates~\cite{angelopoulos2023conformal}. This condition also provides an estimate of the uncertainty associated with selecting the appropriate skill; therefore, the query is forwarded to the LLM for further disambiguation.
In the last case ($iii$), where only one skill $p^*$ exhibits a score lower than its corresponding threshold, this skill is selected as the correct skill to execute in response to the operator's query $q$:
\begin{equation}\label{eq:p_star}
p^*=p_k \text{ s.t. score}(e_q, \mathcal{E}^k)< \lambda_k, \text{ and score}(e_q,\mathcal{E}^j)<\lambda_j, \forall j\neq k 
\end{equation}
This mechanism provides to uncertainty quantification regarding whether the system is able to respond to the user instruction only using a single skill.

\subsubsection{Language Model for Skill Sequence Generation}
\label{subsubsec:llm}
As described in Sec.~\ref{subsubsec:conf_pred}, the conformal prediction module determines whether the input corresponds to a known manipulation skill.
If the classification is uncertain, the instruction is forwarded to the LLM for task decomposition and out-of-distribution detection.

In this case the query $q$ in natural language could either describe a new and unscripted task or a request outside the capabilities of the robot.
An autoregressive generation of an LLM is triggered to evaluate the query and either determine that the requested task is out-of-distribution or compose a long-horizon task by associating sequence $s$ of skills from the available ones $\mathcal{P}=\{p_k\}_{k=1}^K$. 

The value of using an LLM in this context lies in its ability to leverage its pretraining on large internet-scale data to interpret the description provided by the user and the possible implications of the request.

The language model is prompted with the list of available actions, available objects and the specific parameters each action can take as input. The prompt contains a set of examples in a few-shot manner to steer the generation toward a correct output format. It relies on a chain-of-thought structure to generate either a sequence of actions $s=[p_1,p_2,\dots,p_n]$ or a predefined token if the request cannot be fulfilled by the robot.
A constrained decoding is applied to reduce the risk of hallucination and ensure the format of the generation is correct i.e. during sampling, generated skills that do not agree with the set of available skills are masked.

\section{Design of Experiments}
\label{sec:design_exp}

The experimental campaign is designed to answer the following research questions:
\begin{itemize}
    \item \textbf{RQ1:} Can DROM learn robust manipulation skills from a limited number of human demonstrations through DMP-based trajectory augmentation?
    \item \textbf{RQ2:} Can a unified diffusion model effectively encode multiple manipulation skills—including orientation-sensitive behaviors that are difficult to hard-code—and generalize them beyond the demonstrated workspace?
    \item \textbf{RQ3:} Can the learned skills be composed from natural language instructions to solve long-horizon robotic manipulation tasks?
\end{itemize}

To answer these research questions, we first introduce the manipulation skill library (Sec.~\ref{subsec:skills}) and describe the dataset generation procedure (Sec.~\ref{subsec:traj_gen_details}), addressing RQ1 through the collection and DMP-based augmentation of expert demonstrations.
We then present the implementation details of the diffusion model (Sec.~\ref{subsec:DM_details}), which serves as the core trajectory generator and is evaluated to answer RQ2.
Finally, we describe the language-guided manipulation pipeline, including the language encoder (Sec.~\ref{subsec:encoder_details}), the language model (Sec.~\ref{subsec:LLM_details}), and the long-horizon manipulation tasks (Sec.~\ref{subsec:tasks}), which together enable skill selection and long-horizon task decomposition, thereby addressing RQ3.
Finally, Sec.~\ref{subsec:impl_details} highlights the implementation details of the DROM framework.

\subsection{Skill Description}
\label{subsec:skills}

\begin{figure}[ht]
    \centering
    \includegraphics[width=1.0\linewidth]{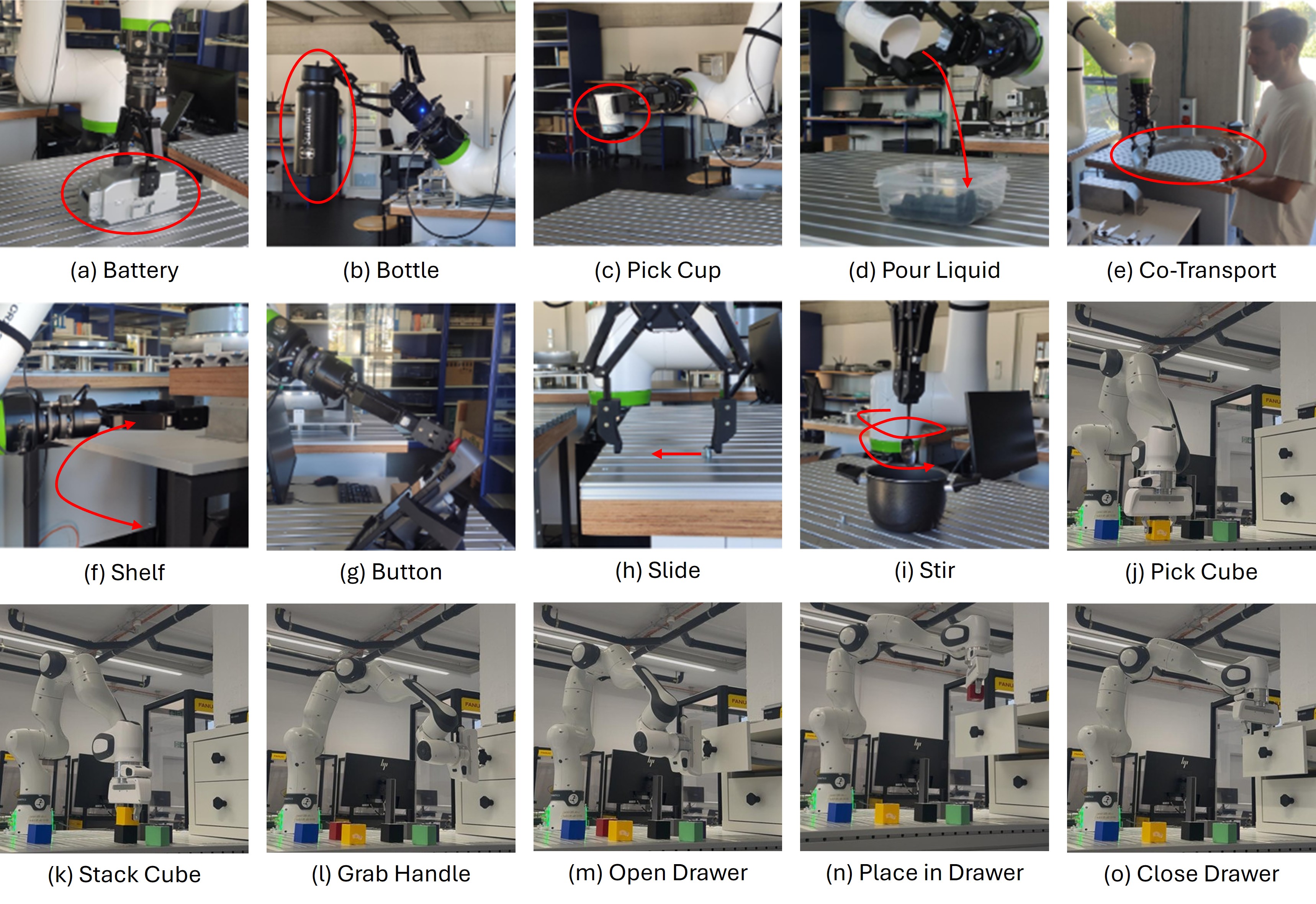}
    \caption{The set of manipulation skills considered in this study for the FANUC CRX25ia robot (a–i) and the Franka Emika Panda (j–o).}
    \label{fig:tasks}
\end{figure}

To evaluate the performances of the proposed method, we selected $17$ different primitives from two different robotic arms, namely a Franka Emika Panda and a FANUC CRX25ia, as depicted in Fig.~\ref{fig:tasks}.
Each skill has specific constraints, for which we define dedicated success metrics.

\begin{itemize}

    \item[-] \textbf{Pick Battery:} Grasp an e-bike battery from the workspace and lift it to a predefined height. Success requires a stable grasp without collisions and lifting the battery by at least $10$~cm above its initial position.

    \item[-] \textbf{Place Battery:} Transport the grasped battery to a designated location while maintaining a stable grasp throughout the motion. Success requires placing the battery within $0.5$~cm of the target pose without dropping or tilting it.

    \item[-] \textbf{Pick Bottle:} Grasp a bottle by inserting one gripper finger into its handle, requiring accurate end-effector alignment before lifting. Success requires complete collision-free insertion of the gripper finger into the handle, and lifting the bottle by at least $10$~cm.

    \item[-] \textbf{Place Bottle:} Transport the lifted bottle to a target location while maintaining the finger engaged inside the handle. Success requires placing the bottle upright within $0.5$~cm of the target position and safely extracting the gripper without displacing the bottle or contacting the table.

    \item[-] \textbf{Pick Cup:} Grasp a liquid-filled cup while maintaining an appropriate end-effector orientation to minimize liquid oscillation. Success requires securely grasping the cup, lifting it by at least $10$~cm, and preventing visible liquid spillage. The transferred volume is quantified by measuring the weight of the cup before and after the picking operation.

    \item[-] \textbf{Pour Liquid:} Transport the grasped cup and execute a controlled wrist rotation to pour the liquid into a target container. Success requires transferring at least $95\%$ of the liquid into the target container without visible spillage. The transferred volume is quantified by measuring the weight of the cup before and after the pouring operation.

    \item[-] \textbf{Co-Transport:} Collaboratively transport a heavy object with a human while maintaining a vertical orientation and ergonomic carrying height. Success requires preserving the object orientation within $5^\circ$ and its position within $5$~cm throughout the transport.

    \item[-] \textbf{Shelf:} Navigate the end-effector through a constrained shelf by executing collision-free top-to-bottom and bottom-to-top trajectories. Success requires completing both motions without collisions while remaining within the shelf boundaries.

    \item[-] \textbf{Button:} Approach and press a push button along its activation axis with accurate alignment. Success requires fully activating the button without lateral deviations or unintended contacts.

    \item[-] \textbf{Slide:} Guide a nut along a linear slot while maintaining constant height, alignment, and continuous contact. Success requires completing the traversal of the $20$-cm slot without generating excessive lateral forces.

    \item[-] \textbf{Stir:} Insert a spoon into a container and execute circular stirring motions while maintaining a vertical end-effector orientation. Success requires completing at least five full revolutions without collisions.

    \item[-] \textbf{Pick Cube:} Grasp a cube using a collision-free approach and lift it from the workspace. Success requires a secure grasp and lifting the cube by at least $10$~cm.

    \item[-] \textbf{Stack Cube:} Place the grasped cube on top of another cube with accurate alignment. Success requires stable stacking for at least $5$~s without displacing or toppling either cube.

    \item[-] \textbf{Grab Handle:} Approach and securely grasp a drawer handle with precise alignment. Success requires establishing a stable grasp without collisions and maintaining contact during the subsequent pulling motion.

    \item[-] \textbf{Open Drawer:} Pull the grasped drawer handle to fully open the drawer while maintaining a stable end-effector orientation. Success requires opening the drawer to at least $95\%$ of its maximum travel without collisions or abrupt motions.

    \item[-] \textbf{Place in Drawer:} Place the manipulated object inside the opened drawer with accurate positioning. Success requires depositing the object inside the drawer without contacting its boundaries or displacing the drawer.

    \item[-] \textbf{Close Drawer:} Push the drawer to its fully closed position through a smooth and controlled motion. Success requires completely closing the drawer without collisions or excessive forces.

\end{itemize}

\subsection{Training Dataset Generation Details}
\label{subsec:traj_gen_details}
\begin{table}[ht]
    \centering
    \begin{tabular}{lc@{\hspace{1.5em}}|@{\hspace{1.5em}}lc}
        \toprule
        \textbf{Skill} & \textbf{Demos} &
        \textbf{Skill} & \textbf{Demos} \\
        \midrule
        Co-Transport & 1 & Pick Battery  & 2 \\
        Shelf        & 1 & Place Battery & 2 \\
        Slide        & 1 & Pick Bottle   & 4 \\
        Stir         & 1 & Place Bottle  & 4 \\
        Pick Cube    & 1 & Pick Cup      & 4 \\
        Stack Cube   & 1 & Pour Liquid   & 4 \\
        Open Drawer  & 1 & Button        & 4 \\
        Place in Drawer & 1 & Grab Handle   & 2 \\
        Close Drawer & 1 &               &   \\
        \bottomrule
    \end{tabular}
    \caption{Number of demonstrations collected for each manipulation skill.}
    \label{tab:skill_demos}
\end{table}

The training dataset is constructed by augmenting expert demonstrations collected for each manipulation skill.
Demonstrations are acquired through kinesthetic teaching on the real robotic platforms or via a SpaceMouse interface in simulation, and are recorded at a frequency of $30$~Hz (Sec.~\ref{subsubsec:data_collection}).

The number of demonstrations collected for each skill depends on the complexity of the corresponding manipulation behavior and is summarized in Tab.~\ref{tab:skill_demos}.
In particular, the number of demonstrations is chosen to adequately cover the workspace and capture the variability of the manipulation strategy associated with each skill.
For tasks whose execution remains largely invariant across the workspace, a single demonstration is sufficient to characterize the desired motion and enable DMP-based trajectory augmentation (Tab.~\ref{tab:skill_demos} Left).
Conversely, skills involving orientation-sensitive interactions or requiring precise approach motions necessitate multiple demonstrations to capture the different valid manipulation modes that may arise across the workspace (Tab.~\ref{tab:skill_demos} Right).

A separate Dynamic Movement Primitive (DMP) is then learned for each demonstration (Sec.~\ref{subsubsec:dmps_augmentation}).
The trajectories generated by all DMPs associated with the same skill are combined to produce a fixed dataset of $300$ validated trajectories per skill.
Trajectory generation is performed by perturbing the initial and goal configurations while preserving the demonstrated motion pattern.
The resulting trajectories are replayed, validated according to task-specific success criteria, resampled to a fixed horizon $H$, and paired with the corresponding language embedding to obtain the final training dataset.

The training workspace boundaries for object placement are defined as follows:
for the cube manipulated with the Franka robot, $P_x \in [+0.2, +0.5]$, $P_y \in [-0.1, +0.1]$, $P_z \in [0.0, +0.05]$, and $R_z \in [0.0, +180]$;
for objects manipulated with the FANUC robot, $P_x \in [+0.3, +0.7]$, $P_y \in [+0.3, +0.45]$, $P_z \in [0.0, 0.05]$, and $R_z \in [-30, +30]$.
Position boundaries are expressed in meters, while orientation ones in degrees.
For skills with relatively simple constraints, such as \textit{Shelf}, \textit{Slide}, and \textit{Stir}, start and goal positions are randomly sampled within the predefined region.

\subsection{Diffusion Model Architecture}
\label{subsec:DM_details}

\begin{figure}[ht]
    \centering
    \includegraphics[width=1\linewidth]{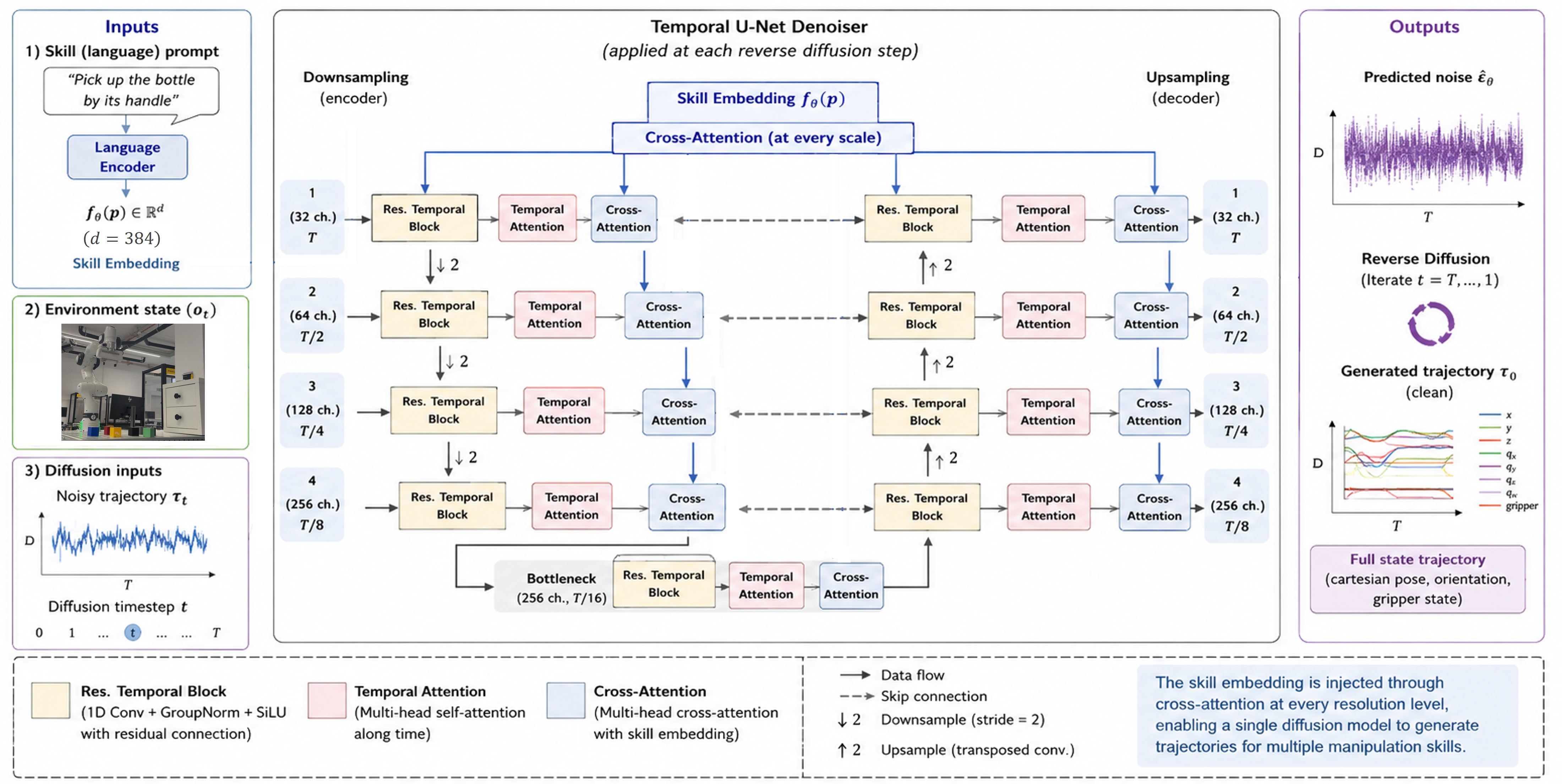}
    \caption{\textbf{Architecture of the proposed skill-conditioned diffusion model.} At each reverse diffusion step, a Temporal U-Net denoiser predicts the noise of the trajectory conditioned on the diffusion timestep, the environment state, and the language embedding of the user prompt. Cross-attention layers at every level of the U-Net allow the model to attend to skill-specific semantic information while preserving temporal structure through the U-Net backbone.}
    \label{fig:DM_architecture}
\end{figure}

For trajectory modeling, we adopt the Temporal U-Net architecture proposed by Carvalho et al.~\cite{carvalho2025motion}, which has demonstrated strong performance in trajectory generation for robotic manipulation.
The network models complete manipulation trajectories rather than individual actions, enabling the diffusion process to capture long-range temporal dependencies and the global structure of motions, including approach, interaction, and retreat phases.

The architecture, depicted in Fig. \ref{fig:DM_architecture}, consists of four temporal downsampling layers with progressively increasing channel dimensions from $32$ to $256$, a bottleneck layer that encodes global temporal information, and a symmetric upsampling path with skip connections that preserve fine-grained motion details during reconstruction.
Each resolution level is composed of Residual Temporal Blocks and temporal attention mechanisms, resulting in a model containing approximately $60$ million trainable parameters.

Different from MPD~\cite{carvalho2025motion}, which is formulated as a single-task planner, we extend the diffusion architecture to support multi-skill trajectory generation through explicit skill conditioning.
Specifically, a learned language embedding $f_\theta(p)$ is injected into the denoising network through a cross-attention mechanism operating at each reverse diffusion step.
This modification allows the denoising process to attend to skill-specific semantic information while preserving the temporal structure of the trajectory.

Consequently, a single diffusion model can represent multiple manipulation primitives and generate skill-consistent trajectories conditioned on the intended task.
The resulting formulation combines trajectory-level generative modeling with semantic conditioning, enabling the model to disambiguate behaviors that may share similar geometric configurations but require fundamentally different motion strategies and object interactions.

\subsection{Language Encoder}
\label{subsec:encoder_details}
The language encoder maps natural language instructions into a compact semantic embedding space in which commands with similar meanings are located close to each other, enabling robust similarity-based retrieval of manipulation skills despite variations in wording.

Compared with larger language encoders, it provides meaningful sentence embeddings while significantly reducing memory consumption and inference time, allowing users to interact with the robot through natural language with minimal latency.

After the embedding of instructions, a similarity-based classification (Sec.~\ref{subsubsec:conf_pred}) determines the target primitive, guiding the diffusion model toward generating a trajectory consistent with the operator’s request.
Results are tested on a set of $600$ utterances annotated with the correspondent primitive.
The dataset contains several variants of verbal requests to generate each skill.
The six skills considered for the evaluation are \emph{Pick Cube}, \emph{Stack Cube}, \emph{Grab Handle}, \emph{Open Drawer}, \emph{Place in Drawer},  and \emph{Close Drawer}.

\subsection{Language Model}
\label{subsec:LLM_details}



To assess linguistic robustness, the benchmark includes instructions with different wording, sentence structures, and levels of complexity, ranging from direct imperative commands (e.g., ``stack the red cube on the blue one'') to more abstract objectives requiring implicit reasoning about the required intermediate manipulation steps (e.g., ``tidy up the workspace by storing all objects in the drawer'').

Since the same task can often be accomplished through different yet equally valid orders of execution, multiple ground-truth plans are defined whenever appropriate.
For example, when cleaning multiple objects from the workspace, different pick-and-place orders are considered correct provided that all task constraints are satisfied.

Performance is quantified using the planning success rate, defined as the percentage of instructions for which the generated skill sequence matches one of the valid ground-truth plans.
A generated plan is considered successful only if it selects the correct manipulation primitives, associates them with the appropriate target objects, and preserves a logically consistent execution order that achieves the requested long-horizon objective.

\subsection{Task Description}
\label{subsec:tasks}

The long-horizon tasks considered in this work are summarized below.

\begin{itemize}

    \item[-] \textbf{Pick and Place}: the robot grasps and relocates different objects, including an e-bike battery and a bottle. This task evaluates the ability of DROM to compose object-specific manipulation skills while adapting the generated trajectories to different object geometries and grasping constraints. Success requires executing the sequence
    $s=[\text{PickObject},\ \text{PlaceObject}]$.
    
    \item[-] \textbf{Pour}: the robot grasps a liquid-filled cup and pours its contents into a target container while maintaining a stable orientation throughout the transport phase. This task evaluates orientation-sensitive manipulation and coordinated translation–rotation motions. Success requires executing the sequence $s=[\text{PickCup},\ \text{PourLiquid}]$.
    
    \item[-] \textbf{Stack}: the robot grasps a cube and accurately places it on top of another cube. This task assesses the composition of grasping and precise placement skills requiring accurate alignment. Success requires executing the sequence $s=[\text{PickCube},\ \text{StackCube}]$.
    
    \item[-] \textbf{Hide}: the robot opens a drawer, stores a cube inside, and closes the drawer. This task evaluates the composition of object manipulation and drawer interaction primitives. Success requires executing the sequence $s=[\text{GrabHandle},\ \text{OpenDrawer},\ \text{PickCube},\ \text{PlaceInDrawer},\ \text{CloseDrawer}]$.
    
    \item[-] \textbf{Clean Up}: the robot removes all cubes from the workspace and stores them inside a drawer. Unlike the previous task, the number of objects is unknown at planning time and varies from one to three cubes, requiring the language model to generate variable-length skill sequences. This represents the most challenging long-horizon scenario, jointly evaluating high-level reasoning and sequential trajectory generation. Success requires executing the sequence $s=[\text{GrabHandle},\ \text{OpenDrawer},\ (\text{PickCube},\ \text{PlaceInDrawer})^{N},\ \text{CloseDrawer}]$,
    where $N\in\{1,2,3\}$ denotes the number of cubes present on the table.

\end{itemize}

Collectively, these tasks provide a comprehensive benchmark for evaluating DROM, allowing us to assess its ability to learn and generalize individual manipulation behaviors, generate skill-consistent trajectories for diverse objects, and compose multiple primitives to achieve complex long-horizon objectives from high-level user instructions.

\subsection{Implementation Details}
\label{subsec:impl_details}

The diffusion model is trained for $500$ epochs using the DDIM scheduler with $25$ denoising steps.
Training is performed with the Adam optimizer, a batch size of $32$ trajectories, and an initial learning rate of $10^{-5}$, with early stopping based on the validation loss.
An exponential variance schedule is adopted, and the model predicts complete manipulation trajectories of fixed horizon $H=64$ rather than individual actions.

Natural language instructions are embedded using the pretrained transformer \verb|all-MiniLM-L6-v2|~\cite{wang2020minilmdeepselfattentiondistillation}, a lightweight distilled BERT-based encoder~\cite{devlin2019bertpretrainingdeepbidirectional} comprising six transformer layers and a $384$-dimensional embedding space.
For long-horizon task planning, we employ \verb|gpt-oss-120b|~\cite{openai2025gptoss120bgptoss20bmodel}, which decomposes high-level user requests into executable manipulation skill sequences.
The language model is evaluated on a benchmark of $50$ natural language instructions for each task category introduced in Sec.~\ref{subsec:tasks}, where each instruction is manually annotated with one or more valid skill sequences.

The proposed framework is implemented in PyTorch and evaluated on a workstation equipped with an NVIDIA RTX 4070 GPU.
Experimental validation is conducted both in MuJoCo simulation and on two collaborative robotic platforms: a FANUC CRX25ia~\cite{franceschi2025ros2fanucinterfacedesign} and a Franka Emika Panda (Fig.~\ref{fig:model_schema}D).

The language encoder processes operator queries in $13.74\pm 15.231$ ms, enabling responsive human--robot interaction.
The language model requires $3996 \pm 2388$ ms per inference (750-run average); however, it is invoked only during high-level task planning and therefore does not affect the real-time control loop.
Training the diffusion model requires approximately $240$ minutes, while inference generates a batch of $10$ candidate trajectories in $281 \pm 427$ ms, enabling efficient online trajectory generation.

Object localization is performed using $50$ mm $\times$ $50$ mm ArUco fiducial markers (\texttt{DICT\_4X4}) attached to each manipulation object.
Marker poses are estimated online using a Perspective-n-Point (PnP) algorithm and transformed into the robot base frame through offline camera calibration.
To improve robustness, the estimated poses are filtered using a $2$ s rolling-window median filter before being incorporated into the environment state for both DMP-based data augmentation and diffusion-based trajectory generation.
The use of fiducial markers allows the evaluation to focus on the proposed manipulation framework while minimizing the influence of perception errors.

\section{Experimental Results}
\label{sec:results}

\subsection{RQ1: Learning Manipulation Skills from Limited Demonstrations}
\label{subsec:rq1}
The goal of this section is to demonstrate that a small number of demonstrations, together with DMP augmentation, is sufficient for training.

\subsubsection{Effect of the number of demonstrations}
\begin{table}[ht]
    \centering
    \begin{tabular}{lccc}
        \toprule
        \textbf{Skill} & \textbf{1 Demo} & \textbf{2 Demos} & \textbf{4 Demos} \\
        \midrule
        Pick Battery     & $48\%$ & $\mathbf{92\%}$ & $\mathbf{92\%}$ \\
        Place Battery     & $50\%$ & $\mathbf{98\%}$ & $\mathbf{98\%}$ \\
        Pick Bottle      & $24\%$ & $72\%$ & $\mathbf{80\%}$ \\
        Place Bottle     & $24\%$ & $78\%$ & $\mathbf{84\%}$ \\
        Pick Cup    & $28\%$ & $46\%$ & $\mathbf{86\%}$ \\
        Pour Liquid & $22\%$ & $56\%$ & $\mathbf{92\%}$ \\
        Button      & $30\%$ & $62\%$ & $\mathbf{82\%}$ \\
        Grab Handle & $60\%$ & $\mathbf{90\%}$ & $\mathbf{90\%}$ \\
        \bottomrule
    \end{tabular}
    \caption{\textbf{Effect of the number of demonstrations on Diffusion Model performance.} Success rates obtained for orientation-sensitive manipulation skills when increasing the number of demonstrations provided to the DMP augmentation process.}
    \label{tab:results_num_demos}
\end{table}

The objective of this study is to evaluate whether multiple demonstrations enable the diffusion model to capture the multimodal nature of orientation-sensitive manipulation skills.
To isolate the effect of demonstration diversity, the diffusion model is always trained on a fixed dataset of $300$ trajectories per skill, while only the number of expert demonstrations used for DMP-based augmentation is varied.
Each model is trained for $500$ epochs and evaluated over $50$ validation trials.

The results are reported in Tab.~\ref{tab:results_num_demos}.
As expected, increasing the number of demonstrations consistently improves performance.
The largest gains are observed for \textit{Pick Bottle}, \textit{Pick Cup}, and \textit{Pour Liquid}, whose success rates increase from $24\%$, $28\%$, and $22\%$ with a single demonstration to $84\%$, $86\%$, and $92\%$, respectively, with four demonstrations.
These skills require different end-effector orientations and approach strategies depending on the object pose, making them inherently multimodal.

The \textit{Button} skill exhibits a similar trend, improving from $30\%$ to $82\%$.
Conversely, the \textit{Battery} and \textit{Grab Handle} skills reach near-optimal performance with only two demonstrations, improving from $48\%$ to $98\%$ and from $60\%$ to $90\%$, respectively.
For the battery, two demonstrations capture the required grasping orientations, whereas for \textit{Grab Handle}, the two demonstrations correspond to the two drawer-handle geometries considered in the experimental setup, after which performance saturates.

Overall, the results demonstrate that increasing demonstration diversity, rather than dataset size, enables the diffusion model to learn a multimodal distribution of valid manipulation strategies and to select the appropriate trajectory according to the object configuration at inference time.

\subsubsection{Generalization outside demonstrations}
\begin{table}[ht]
    \centering
            \begin{tabular}{l|c|c|c|c}
                \toprule
            \textbf{Object} & \multicolumn{1}{c|}{$\mathbf{P_x}$} & \multicolumn{1}{c|}{$\mathbf{P_y}$} & \multicolumn{1}{c|}{$\mathbf{P_z}$} & \multicolumn{1}{c}{$\mathbf{R_z}$} \\
            \midrule
            Battery              & $23\%$ & $21\%$ & $7\%$  & $16\%$ \\
            Bottle               & $24\%$ & $24\%$ & $8\%$  & $9\%$  \\
            Button               & $24\%$ & $24\%$ & $8\%$  & $5\%$  \\
            Co-Transport Object  & $28\%$ & $22\%$ & $7\%$  & $23\%$ \\
            Pouring Container    & $31\%$ & $19\%$ & $3\%$  & $19\%$ \\
            Shelf                & $17\%$ & $29\%$ & $2\%$  & $33\%$ \\
            Slide Nut            & $16\%$ & $26\%$ & $0\%$  & $30\%$ \\
            Stirring Spoon       & $12\%$ & $38\%$ & $5\%$  & $0\%$  \\
            Franka Cube          & $11\%$ & $24\%$ & $14\%$ & $50\%$ \\
            \bottomrule
        \end{tabular}%
    \caption{\textbf{Workspace} of each manipulated object, reported in terms of translational and rotational variations.}
    \label{tab:ws_objects}
\end{table}

The objective of this study is to evaluate the generalization capability of the diffusion model beyond the workspace covered by the expert demonstrations.
Tab.~\ref{tab:ws_objects} reports the maximum workspace expansion, expressed as the percentage increase with respect to the training workspace, for which the generated trajectories remain successful.

Trajectories were trained within the workspace defined in Sec.~\ref{subsec:traj_gen_details} and evaluated over progressively expanded translational and rotational ranges.
Overall, our model generalizes consistently outside the demonstrated region, achieving average workspace expansions of $20.7\%$ and $25.2\%$ along the $X$ and $Y$ axes, respectively.
Generalization along the vertical direction is naturally more limited (average $6.0\%$), as most manipulation skills focus exclusively on objects resting on planar surfaces and therefore exhibit little variability in height during training.

The results also demonstrate that the diffusion model extrapolates to unseen object orientations.
The largest rotational improvements are obtained for the \textit{Franka Cube} ($50\%$), \textit{Shelf} ($33\%$), and \textit{Slide Nut} ($30\%$), whose manipulation strategies remain largely invariant under rotations about the vertical axis.
Conversely, orientation-sensitive skills such as \textit{Bottle}, \textit{Button}, and \textit{Pouring Container} achieve more moderate rotational expansions ($5$--$19\%$), since successful execution requires maintaining specific end-effector orientations and approach directions with respect to the manipulated object.

Finally, the \textit{Stirring Spoon} and \textit{Slide Nut} skills were not evaluated under rotational perturbations because of their rotational symmetry, while the \textit{Slide Nut} was not tested for increased $Z$ positions since its motion is constrained by the table surface.

Overall, these results indicate that the proposed diffusion model learns manipulation behaviors rather than memorizing demonstrated trajectories, enabling reliable extrapolation to unseen object positions and orientations while preserving the skill-specific constraints required for successful execution.

\subsection{RQ2: Multi-Skill Diffusion Policy}
\label{subsec:rq2}
In this section we would like to answer the following research question: ``Can one diffusion model learn many manipulation skills?"

\subsubsection{Imitation Learning Performance}
\begin{table}[ht]
    \centering
    \begin{tabular}{lcc|lcc}
        \toprule
        \multicolumn{3}{c|}{\textbf{FANUC CRX25ia}} &
        \multicolumn{3}{c}{\textbf{Franka Research 3}} \\
        \cmidrule(r){1-3}\cmidrule(l){4-6}
        \textbf{Skill} & \textbf{MPD} & \textbf{DROM} &
        \textbf{Skill} & \textbf{MPD} & \textbf{DROM} \\
        \midrule
        Pick Battery    & $76\%$ & $98\%$ & Pick Cube       & $78\%$ & $90\%$ \\
        Place Battery   & $82\%$ & $96\%$ & Stack Cube      & $44\%$ & $94\%$ \\
        Pick Bottle     & $60\%$ & $84\%$ & Grab Handle     & $38\%$ & $90\%$ \\
        Place Bottle    & $76\%$ & $94\%$ & Open Drawer     & $50\%$ & $86\%$ \\
        Pick Cup        & $38\%$ & $86\%$ & Place in Drawer & $62\%$ & $92\%$ \\
        Pour Liquid     & $52\%$ & $92\%$ & Close Drawer    & $68\%$ & $88\%$ \\
        Button          & $74\%$ & $82\%$ &                 &         &       \\
        Co-Transport    & $42\%$ & $92\%$ &                 &         &       \\
        Shelf           & $66\%$ & $96\%$ &                 &         &       \\
        Slide           & $70\%$ & $94\%$ &                 &         &       \\
        Stir            & $64\%$ & $88\%$ &                 &         &       \\
        \bottomrule
    \end{tabular}%
    \caption{\textbf{Comparison between DROM and MPD} on the considered manipulation skills. The left half reports results on the FANUC CRX25ia, while the right half reports results on the Franka Emika Panda. DROM consistently outperforms MPD across both robotic platforms by learning multiple manipulation skills within a single language-conditioned diffusion model.}
    \label{tab:results_robots}
\end{table}

In this evaluation we benchmark DROM against Motion Planning Diffusion (MPD)~\cite{carvalho2025motion} to evaluate the effect of the introduced architectural extensions on imitation accuracy and skill-consistent trajectory generation.

As described in Sec.~\ref{subsec:DM_details}, training separate models enables a controlled evaluation across robotic platforms with different kinematic structures, workspace dimensions, and manipulation constraints.
At the same time, employing the same network architecture, hyper-parameters, and training pipeline on both platforms ensures a fair comparison and provides a consistent assessment of the scalability and hardware-specific generalization capabilities of the proposed framework.

\begin{figure}[t]
    \centering
    \begin{subfigure}{\linewidth}
        \centering
        \includegraphics[width=\linewidth]{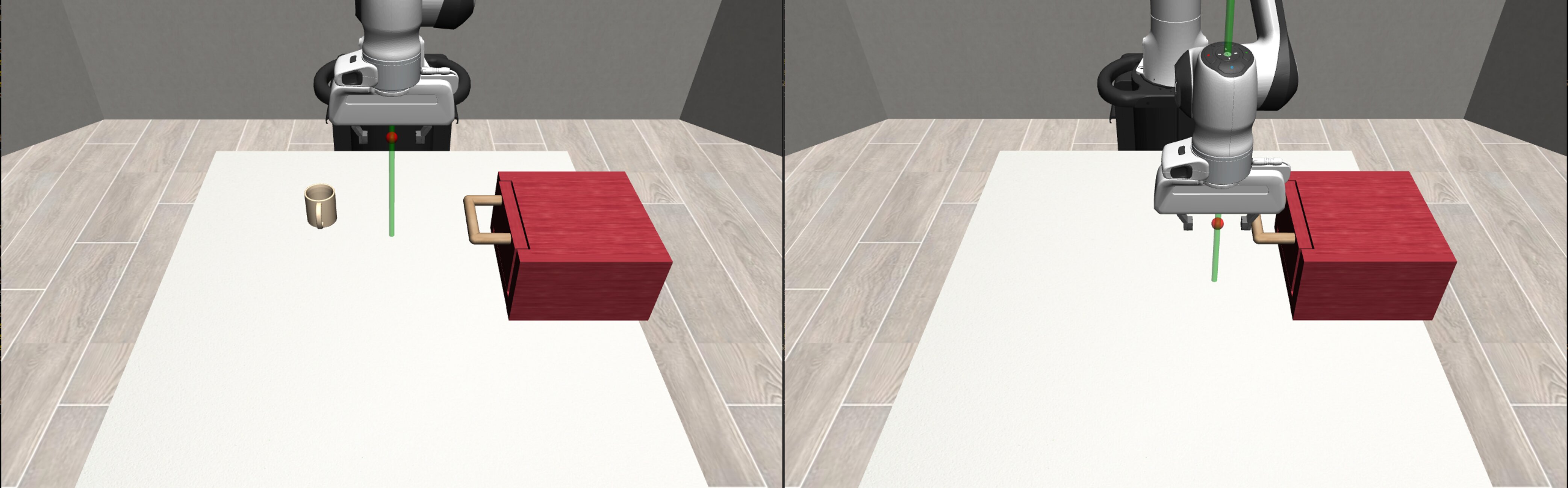}
        \caption{Task 1: Table Clean Up (MugCleanup)}
    \end{subfigure}
    \vspace{0.5em}
    \begin{subfigure}{\linewidth}
        \centering
        \includegraphics[width=\linewidth]{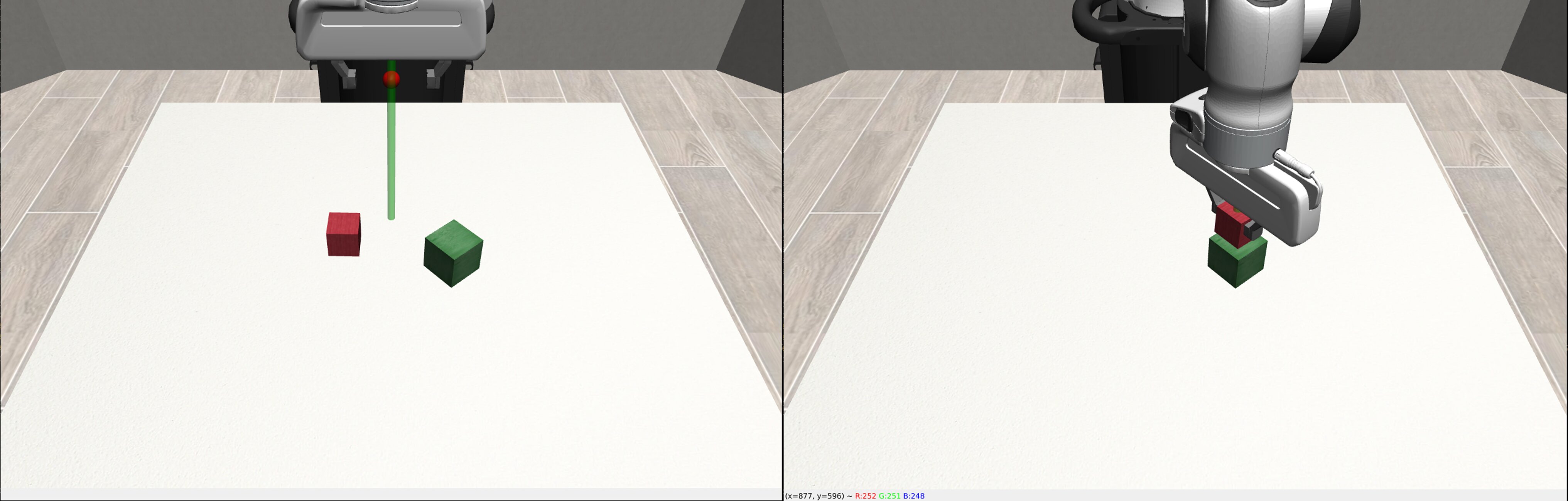}
        \caption{Task 2: Cube stacking (Stack)}
    \end{subfigure}
    \caption{\textbf{Simulation environments.} The two environments are designed to closely replicate the manipulation setups executed on the real Franka arm, while introducing structured, multi-stage objectives that decompose naturally into sequential subtasks.}
    \label{fig:sim_tasks}
\end{figure}

Each model is trained for $500$ epochs and evaluated over $50$ validation trials per skill.
During evaluation, manipulated objects are repeatedly placed at different positions within and beyond the DMP-generated training workspace to assess both execution robustness and spatial generalization.
The quantitative results, reported in Tab.~\ref{tab:results_robots}, show that DROM consistently outperforms MPD across all considered manipulation skills.

The observed performance gains can be primarily attributed to DROM's explicit conditioning on manipulation primitives, as described in Sec.~\ref{subsec:DM_details}, which guides the diffusion process toward generating coherent, goal-oriented, and dynamically consistent trajectories.

This advantage is particularly evident for skills requiring complex object interactions, such as \textit{Pick Bottle}, \textit{Pick Cup}, \textit{Pour Liquid}, \textit{Grab Handle}, \textit{Open Drawer}, and \textit{Close Drawer}.
These tasks demand precise end-effector orientations and highly specific approach strategies.
For example, bottle manipulation requires accurately inserting the fingers into the handle, cup manipulation requires selecting an appropriate grasping side and maintaining a suitable orientation during pouring, and drawer interactions require precise alignment with the two handles prior to contact.
In these scenarios, DROM effectively disambiguates the desired motion pattern and generates skill-compliant trajectories, consistently outperforming MPD.

\subsubsection{Imitation Learning in Simulation}

To further evaluate the effectiveness of DROM, we conduct an extensive simulation campaign using trajectories synthetically generated through the D-MG~\cite{pomponi2026dynamimicgen} pipeline.
This setup enables us to assess DROM’s capability to learn and extract meaningful motion features directly from the structure of the generated trajectory dataset, without relying on additional supervision.

As in the real-robot experiments, we compare our method to MPD~\cite{carvalho2025motion} in MuJoCo using the \textit{Stack} and \textit{Mug Cleanup} Robosuite environments~\cite{Zhu2020robosuiteAM} (Fig.~\ref{fig:sim_tasks}), which mirror real Franka setups and include structured, multi-stage subtasks.

\begin{table}[ht]
    \centering
        \begin{tabular}{l l c c c}
            \toprule
            \textbf{Env.} & \textbf{Skill} & \textbf{MPD} & \textbf{BC} & \textbf{DROM} \\
            \midrule
            \multirow[t]{2}{*}{MugCleanup}
                & Pick Cube  & $59.33 \pm 2.49$ & $96.00 \pm 2.00$ & $97.33 \pm 1.15$ \\
                & Stack Cube & $60.67 \pm 0.94$ & $92.67 \pm 2.31$ & $94.67 \pm 0.94$ \\
            \cmidrule(lr){1-5}
            \multirow[t]{5}{*}{Stack}
                & Grab Handle     & $28.00 \pm 1.63$ & $92.00 \pm 3.46$ & $96.00 \pm 1.63$ \\
                & Open Drawer     & $58.00 \pm 1.63$ & $90.67 \pm 0.94$ & $96.67 \pm 0.94$ \\
                & Pick Mug        & $28.00 \pm 2.83$ & $85.33 \pm 4.11$ & $88.00 \pm 1.63$ \\
                & Place in Drawer & $30.67 \pm 0.94$ & $92.00 \pm 1.63$ & $94.67 \pm 2.49$ \\
                & Close Drawer    & $44.67 \pm 3.40$ & $91.33 \pm 1.89$ & $95.33 \pm 3.77$ \\
            \bottomrule
        \end{tabular}
    \caption{\textbf{Comparison} of DROM, MPD, and BC, in the MuJoCo simulation environments: Stack and Mug Cleanup.}
    \label{tab:results_sim}
\end{table}

As reported in Tab.~\ref{tab:results_sim}, policies are evaluated using $50$ rollouts per agent checkpoint during training, and we report the average success rate and relative standard deviation across three random seeds.
The Diffusion Model and MPD are trained with the hyperparameters described in Sec.~\ref{subsec:DM_details}, while Behavior Cloning follows the settings in~\cite{pomponi2026dynamimicgen}.
Across all evaluated skills and simulation environments, DROM consistently achieves higher success rates than MPD and Behavior Cloning (BC) baselines.

The performance gap between DROM and MPD is both substantial and consistent, indicating that explicit skill conditioning within the diffusion process markedly improves trajectory quality and execution robustness, leading to smoother approaches and more accurate target alignment.

\begin{figure}[t]
    \centering
    \includegraphics[width=1.0\linewidth]{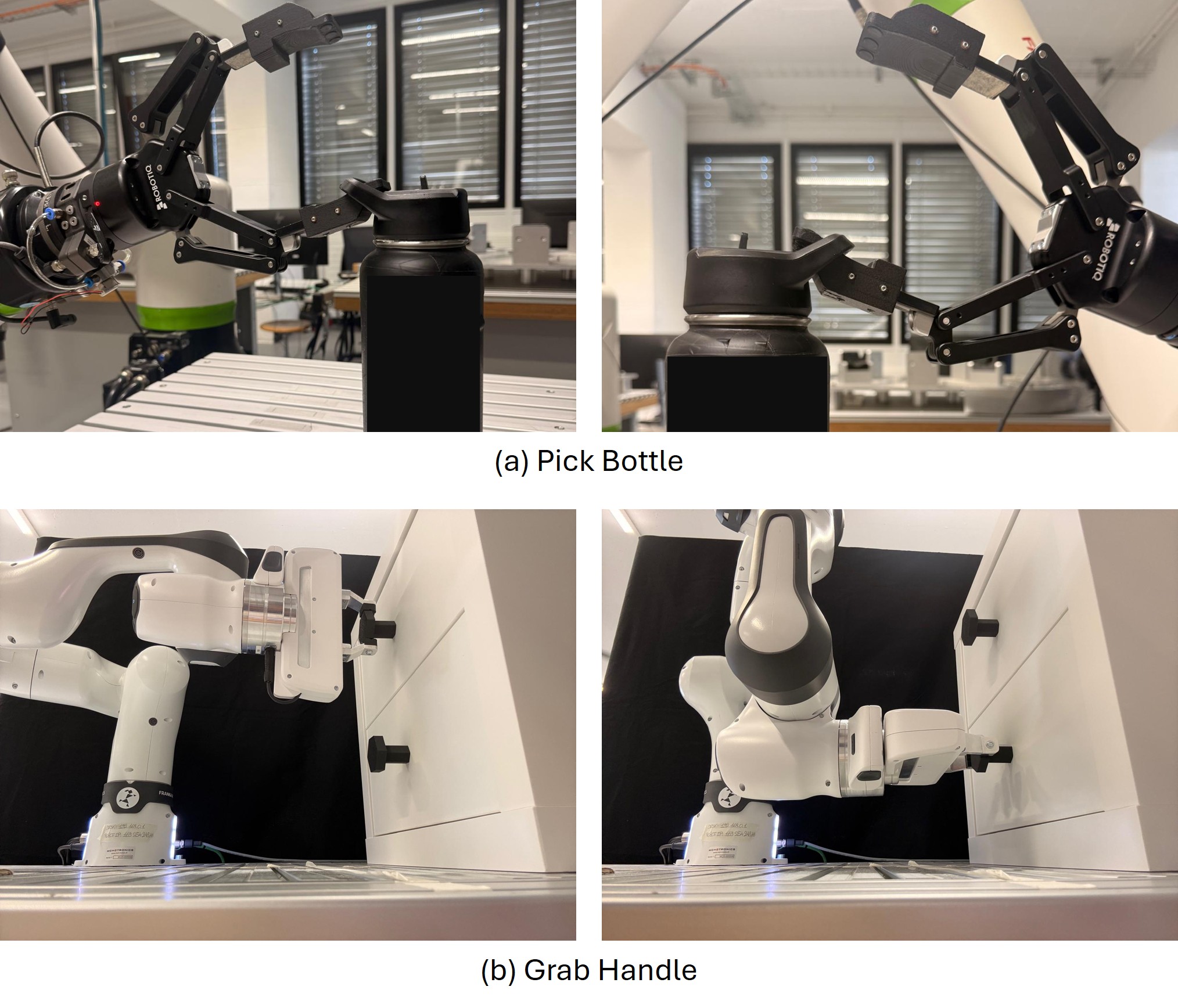}
    \caption{\textbf{Qualitative examples of multimodal trajectory generation.} (a) \textit{Pick Bottle}: depending on the bottle orientation, the diffusion model generates different approach trajectories to insert a gripper finger into the handle. (b) \textit{Grab Handle}: the model adapts the grasping motion to two drawers with different handle geometries, producing the corresponding manipulation strategy.}
    \label{fig:rep_skills}
\end{figure}

Furthermore, DROM consistently outperforms the Behavior Cloning (BC) baseline, highlighting the advantages of diffusion-based imitation learning.
Unlike BC, which predicts actions in a receding-horizon manner and focuses on local control decisions, the diffusion model generates the entire trajectory distribution, explicitly encoding global temporal structure.
This holistic representation allows the policy to account for both past and future motion context at each timestep, resulting in more coherent plans, and higher success rates.

\subsubsection{Diffusion Model Multi-Modality}

The objective of this experiment is to qualitatively demonstrate that the proposed multi-skill diffusion model learns a multimodal distribution of manipulation behaviors rather than memorizing a single trajectory for each skill.
Specifically, when trained with demonstrations containing multiple valid execution strategies, the model generates trajectories that are consistent with both the commanded skill and the current object configuration, selecting the appropriate manipulation mode at inference time.

The evaluation is conducted on two representative orientation-sensitive skills, one performed on the FANUC CRX25ia and one on the Franka Emika Panda.
For each skill, the experiment is repeated over $10$ independent trials to assess both the model's ability to identify the correct manipulation mode and the resulting execution success.
Representative examples are shown in Fig.~\ref{fig:rep_skills}.

Figure~\ref{fig:rep_skills}(a) illustrates the \textit{Pick Bottle} skill, where successful grasping requires inserting one gripper finger into the bottle handle.
Since the handle orientation varies with the object pose, the end-effector must approach the bottle from different directions.
The evaluation considers four distinct grasping modes corresponding to different bottle orientations.
Across all $10$ trials, the diffusion model correctly selected the appropriate approach strategy ($10/10$), demonstrating that it successfully disambiguates all four manipulation modes.
The overall task was successfully completed in $8/10$ trials, with the remaining failures caused by inaccuracies during the physical grasping phase rather than by an incorrect mode selection.

Similarly, Fig.~\ref{fig:rep_skills}(b) depicts the \textit{Grab Handle} skill.
The experimental setup includes two drawers with identical handle geometries but different orientations, each requiring a different approach direction and grasp configuration.
The diffusion model correctly identified the appropriate manipulation mode in all $10$ trials and successfully completed the task in $9/10$ executions.
The only failure was due to the interaction with the drawer handle rather than an incorrect selection of the approach strategy.

These qualitative results complement the quantitative analysis presented in Sec.~\ref{subsec:rq1} by showing that the diffusion model captures the underlying distribution of valid manipulation strategies rather than individual demonstrations.
While the execution success is ultimately affected by the physical interaction with the environment, the model consistently infers the correct manipulation mode according to the observed object configuration.

\subsection{RQ3: Language-Guided Diffusion Manipulation}
\label{subsec:rq3}

This subsection addresses \textbf{RQ3} by evaluating DROM's capability to solve manipulation tasks directly from natural language instructions.
Specifically, we assess the complete language-guided manipulation pipeline, from interpreting the operator's request to generating and executing the corresponding sequence of manipulation skills.
To this end, we first evaluate the overall performance of DROM on representative long-horizon tasks (listed in Sec.~\ref{subsec:lang_overall}).
We then perform an ablation study on the language encoder to assess its ability to correctly associate natural language instructions with the corresponding manipulation primitives (Sec.~\ref{subsec:lang_em}).
We do the same on the large language model responsible for decomposing high-level user requests into executable sequences of skills, analyzing its contribution to the overall task execution performance (Sec.~\ref{subsec:lang_llm}).
Finally, we evaluate the uncertainty quantification module quantifying its ability to discriminate between skills and tasks requests and out-of-distribution requests (Sec.~\ref{subsec:lang_cp}).

\subsubsection{DROM Performance}
\label{subsec:lang_overall}
\begin{table}[ht] 
    \centering
    \begin{tabular}{l|c|c|c}
        \toprule
        \textbf{Task} &
        \textbf{\# Skills} &
        \textbf{\# Objects} &
        \textbf{SR} \\
        \midrule
        Pick and Place & 2 & 2 & $86.0\%$ \\
        Pour           & 2 & 1 & $70.0\%$ \\
        Stack          & 2 & 2 & $67.5\%$ \\
        Hide           & 5 & 1 & $70.4\%$ \\
        Clean Up       & $5$--$9$ & $1$--$3$ & $60.9\%$ \\
        \bottomrule
    \end{tabular}
    \caption{\textbf{End-to-end evaluation of DROM.} Success rate (SR) achieved on long-horizon manipulation tasks of increasing complexity.}
    \label{tab:drom_performance}
\end{table}

DROM is designed to solve long-horizon manipulation tasks by composing previously learned manipulation primitives.
Accordingly, this evaluation assesses the ability of the proposed framework to interpret high-level natural language instructions, identify the required manipulation skills, and execute them as coherent action sequences to accomplish complex objectives.

The quantitative results are summarized in Tab.~\ref{tab:drom_performance}, which reports the end-to-end success rate (SR) over $50$ validation episodes together with the number of manipulation skills and manipulated objects involved in each task. The evaluation also comprises $200$ different utterances as input for the tasks as will be discussed more in detail in Sec.~\ref{subsec:lang_llm}.
The evaluated scenarios progressively increase in complexity, ranging from short tasks composed of two skills to long-horizon tasks requiring up to nine consecutive manipulation primitives.

Among the tasks composed of only two skills, \textit{Pick and Place} achieves the highest success rate of $86.0\%$, followed by \textit{Pour} with $70.0\%$ and \textit{Stack} with $67.5\%$. These tasks require the composition and execution of only two manipulation primitives, resulting in relatively short-horizon behaviors. The lower performance observed for \textit{Pour} and \textit{Stack} indicates that task success depends not only on sequence length but also on the reliability of the individual manipulation primitives involved.

The \textit{Pick and Place} task achieves the highest overall success rate of $86.0\%$ despite involving two manipulation primitives. This performance is consistent with the $100.0\%$ skill-composition success reported in Table~\ref{tab:results_llm}, with the remaining failures arising during sequence execution. In particular, the task includes the bottle manipulation primitive, which requires accurately inserting one gripper finger into the bottle handle while maintaining the appropriate end-effector orientation during transportation and placement.

The \textit{Hide} task, which requires the sequential execution of five manipulation primitives, achieves a success rate of $70.4\%$. Despite its longer horizon, DROM successfully composes and executes a heterogeneous sequence of manipulation skills, including drawer manipulation and object transportation. The result demonstrates that increasing the number of primitives does not necessarily lead to a proportional reduction in task success when the individual skills and their composition can be reliably executed.

The \textit{Clean Up} task represents the most variable manipulation scenario. The robot must remove between one and three objects from the workspace and place them inside a drawer, resulting in a sequence length that varies between five and nine manipulation primitives depending on the observed scene. For example, when two cubes are present, DROM generates the sequence \textit{Grab Handle}, \textit{Open Drawer}, \textit{Pick Cube$_1$}, \textit{Place Cube$_1$ in Drawer}, \textit{Pick Cube$_2$}, \textit{Place Cube$_2$ in Drawer}, and \textit{Close Drawer}.

Successfully completing \textit{Clean Up} therefore requires both correct high-level sequence composition and reliable execution of multiple consecutive manipulation primitives. DROM achieves an end-to-end success rate of $60.9\%$ in this setting. This result is lower than those obtained for \textit{Pick and Place}, \textit{Pour}, and \textit{Hide}, while remaining comparable to the performance observed for \textit{Stack}. Overall, the results show that DROM can execute variable-length manipulation plans involving multiple objects and heterogeneous skills, while longer and more variable sequences introduce additional opportunities for failure.

Overall, the results show that DROM effectively combines high-level language reasoning with diffusion-based trajectory generation to solve long-horizon manipulation tasks.
As expected, the success rate gradually decreases as the number of required skills and manipulated objects increases due to error accumulation across consecutive primitives.
Despite this increased complexity, the framework consistently generates coherent manipulation sequences and successfully executes a broad range of language-guided robotic tasks.

\subsubsection{Ablation Study: LLM Performance}
\label{subsec:lang_llm}
\begin{table}[ht]
    \centering
        \begin{tabular}{l|c|c|c}
            \toprule
            \textbf{Task} & \multicolumn{1}{c|}{\textbf{Skill Composition}} & \multicolumn{1}{c|}{\textbf{Sequence Execution}} & \multicolumn{1}{c}{\textbf{Overall}}\\
            \midrule
            Pick and Place & $100\%$ & $86.0\%$ & $86.0\%$ \\
            Pour        & $87.5\%$ & $80.0\%$ & $70.0\%$ \\
            Stack       & $75.0\%$ & $90.0\%$ & $67.5\%$ \\    
            Hide        & $80.0\%$ & $88.0\%$ & $70.4\%$ \\    
            Clean Up     & $72.5\%$ & $84.0\%$ & $60.9\%$ \\    
            \bottomrule
        \end{tabular}%
    \caption{Success rates of skill composition, sequence execution and overall for Stack and Clean Up tasks.}
    \label{tab:results_llm}
\end{table}
%

The objective of this experiment is to evaluate DROM's ability to solve long-horizon manipulation tasks by combining high-level language reasoning with diffusion-based multi-skill trajectory generation.
Given a natural language instruction, the large language model first decomposes the requested task into an ordered sequence of manipulation primitives, which are subsequently executed by the diffusion model.

Table~\ref{tab:results_llm} reports three complementary performance metrics, averaged among $250$ queries and evaluated on five different seeds.
The \textit{Skill Composition} success rate measures the percentage of user requests that are correctly translated into the intended sequence of manipulation primitives.
The \textit{Sequence Execution} success rate evaluates the capability of the diffusion model to successfully execute the generated sequence assuming that the decomposition is correct.
Finally, the \textit{Overall} success rate measures the end-to-end performance, requiring both a correct task decomposition and the successful execution of every manipulation primitive.

The results show that skill composition varies across tasks, with the highest success rate obtained for \textit{Pick and Place} ($100.0\%$). \textit{Pour} achieves $87.5\%$, followed by \textit{Hide} ($80.0\%$), \textit{Stack} ($75.0\%$), and \textit{Clean Up} ($72.5\%$). The lower composition performance observed for the more complex tasks indicates the difficulty of generating the correct sequence of manipulation primitives from natural language, particularly when multiple skills must be combined. This effect is most evident for \textit{Clean Up}, which requires the composition of a variable number of skills depending on the objects present in the scene.

Conditioned on a correct skill sequence, the diffusion model achieves consistently high execution success across all tasks. The execution success rate ranges from $80.0\%$ for \textit{Pour} to $90.0\%$ for \textit{Stack}. \textit{Pick and Place}, \textit{Hide}, and \textit{Clean Up} achieve $86.0\%$, $88.0\%$, and $84.0\%$, respectively. These results indicate that, once the high-level sequence is correctly specified, the diffusion-based controller can reliably execute the resulting multi-skill manipulation sequence. The lower execution rate for \textit{Pour} is consistent with the orientation-sensitive nature of the pouring motion, while the execution of \textit{Pick and Place} can be affected by the bottle manipulation primitive, which requires precise insertion of a gripper finger into the bottle handle.

The overall success rate reflects the combined effect of skill composition and sequence execution. \textit{Pick and Place} achieves the highest end-to-end success rate at $86.0\%$, followed by \textit{Hide} ($70.4\%$), \textit{Pour} ($70.0\%$), \textit{Stack} ($67.5\%$), and \textit{Clean Up} ($60.9\%$). In particular, the overall rates are consistent with the product of the corresponding composition and execution rates, indicating that failures at either stage directly reduce end-to-end task success. For \textit{Clean Up}, the combination of the lowest composition rate and a relatively long, variable-length sequence results in the lowest overall performance.

Overall, these results demonstrate that the proposed hierarchical architecture effectively combines language reasoning with diffusion-based trajectory generation for long-horizon robotic manipulation. While the diffusion model executes generated skill sequences with consistently high reliability, the end-to-end performance is primarily limited by the complexity of decomposing variable-length tasks into correct manipulation primitives. These findings suggest that improving high-level task planning, particularly for scenarios involving dynamic numbers of objects and repeated skills, represents the most promising direction for further increasing the overall performance of DROM.

\subsubsection{Failure Analysis}

\begin{table}[ht]
    \centering
    \begin{tabular}{l|l|c|c}
        \toprule
        \textbf{Category} &
        \textbf{Failure Mode} &
        \textbf{\#} &
        \textbf{\%} \\
        \midrule
        \multirow{1}{*}{Language Model}
        & Incorrect skill sequence & $42/250$ & 16.8 \\
        \midrule
        \multirow{6}{*}{Diffusion Model}
        & Failed grasp & $11/250$ & 4.4 \\
        & Placement failure & $7/250$ & 2.8 \\
        & Collision & $7/250$ & 2.8 \\
        & Unfeasible trajectory & $1/250$ & 0.4 \\
        & Goal not reached & $10/250$ & 4 \\
        \bottomrule
    \end{tabular}
    \caption{\textbf{Failure analysis of DROM.} Failures are grouped according to the corresponding module of the framework.}
    \label{tab:failure_analysis}
\end{table}

Table~\ref{tab:failure_analysis} summarizes the failures observed during the evaluation of the long-horizon manipulation tasks, categorizing them according to the module responsible for the unsuccessful execution.

The largest source of failures originates from the language model, which generates an incorrect skill sequence in $16.8\%$ of the evaluated episodes.
These errors mainly occur for high-level requests admitting multiple valid execution strategies or involving a variable number of manipulated objects, such as the \textit{Clean Up} task.
An incorrect skill composition prevents the framework from accomplishing the requested objective regardless of the quality of the generated trajectories.

Failures attributed to the diffusion model are considerably less frequent and are primarily related to challenging physical interactions.
The most common failure mode is an unsuccessful grasp ($4.4\%$), particularly for orientation-sensitive objects such as the bottle, where accurate end-effector alignment is required before establishing contact.
Similarly, failures to reach the desired goal configuration account for $4.0\%$ of the executions and are generally associated with accumulated tracking errors or slight inaccuracies during long manipulation sequences.

Placement failures and collisions each represent only $2.8\%$ of the evaluated episodes.
These errors mainly occur during contact-rich interactions, such as object insertion into the drawer or bottle placement, where small pose deviations may lead to unsuccessful object release or unintended contacts with the environment.
Finally, unfeasible trajectories are rarely observed ($0.4\%$), indicating that the proposed diffusion model consistently generates smooth and executable motions even for a diverse set of manipulation skills.

Overall, the failure analysis indicates that the principal limitation of DROM is not the generation of manipulation trajectories, but rather the complexity of composing and executing long sequences of interdependent skills.
While the diffusion model exhibits robust trajectory generation, errors occurring during language-guided task decomposition or early manipulation stages propagate through the remaining execution, ultimately reducing the end-to-end success rate of long-horizon tasks.

\subsubsection{Ablation Study: Language Encoder Performance} 
\label{subsec:lang_em}
\begin{table}[ht]
    \centering
    \begin{tabular}{l|c|c|c|c}
        \toprule 
        \textbf{Skill} & \textbf{Precision} & \textbf{Recall} & \textbf{F1-score} & \textbf{SR} \\
        \midrule
        Pick Cube       & $96.5\%$ & $83.0\%$ & $89.2\%$ & $89.0\%$ \\
        Stack Cube      & $97.6\%$ & $82.0\%$ & $89.1\%$ & $85.0\%$ \\
        \midrule
        Grab Handle     & $74.6\%$ & $85.0\%$ & $79.4\%$ & $82.0\%$ \\
        Open Drawer     & $85.6\%$ & $89.0\%$ & $87.3\%$ & $85.0\%$ \\
        Place in Drawer & $75.0\%$ & $93.0\%$ & $83.0\%$ & $93.0\%$ \\
        Close Drawer    & $96.6\%$ & $85.0\%$ & $90.4\%$ & $83.0\%$ \\
        \midrule
        \textbf{Average} & $\mathbf{87.6\%}$ & $\mathbf{86.2\%}$ & $\mathbf{86.4\%}$ & $\mathbf{86.2\%}$ \\
        \bottomrule
    \end{tabular}%
    \caption{\textbf{Language encoder evaluation} on a balanced test set of 600 natural language instructions. Precision, recall, F1-score, and class-wise success rate (SR) quantify the encoder's ability to associate operator requests with the corresponding manipulation primitive.}
    \label{tab:language_encoder}
\end{table}

\begin{figure}
    \begin{center}
    \resizebox{\textwidth}{!}{\input{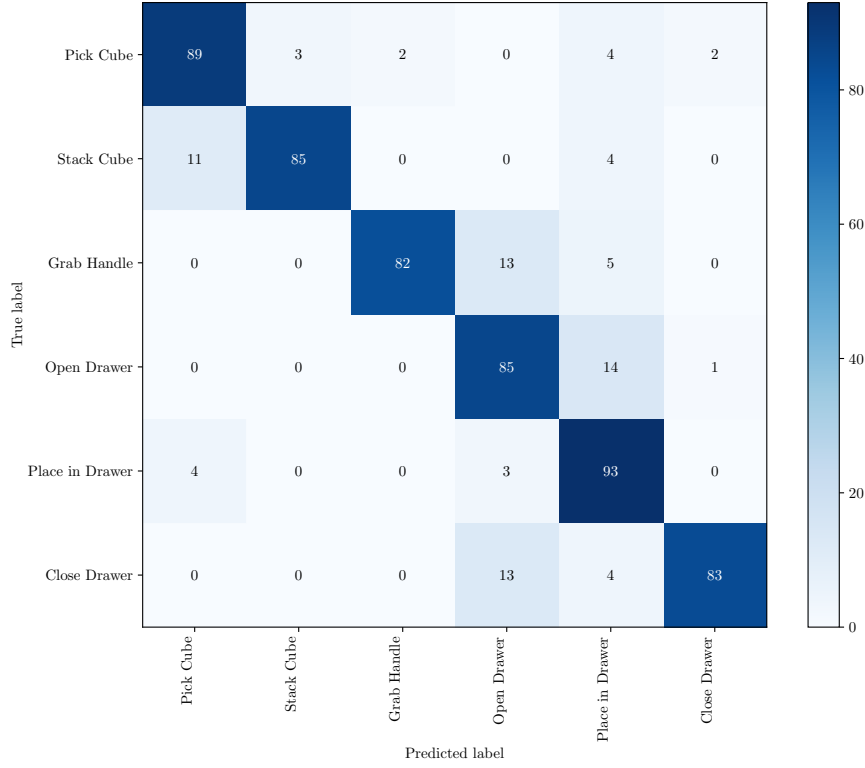}}
    \end{center}
    \caption{\textbf{Confusion matrix} of the language encoder on the six manipulation primitives executed with the Franka Emika Panda robot.}
    \label{fig:confusion_language}
\end{figure}

The language-guided component of DROM is evaluated on the six manipulation primitives (introduced in Sec.~\ref{subsec:skills}) executed with the Franka Emika Panda robot using a balanced dataset of $600$ annotated natural language instructions (100 utterances per primitive).
The evaluation first assesses the ability of the language encoder to associate an operator request with the intended manipulation primitive and subsequently analyzes the end-to-end performance obtained when the recognized skill is executed by the diffusion model.

Table~\ref{tab:language_encoder} reports the classification performance of the language encoder in terms of precision, recall, F1-score, and class-wise success rate (SR).
Overall, the encoder achieves an average precision of $87.6\%$, a recall of $86.2\%$, an F1-score of $86.4\%$ and a success rate of $86.2\%$, demonstrating reliable semantic discrimination among the considered primitives.
The highest f1-scores are achieved for \textit{Pick Cube}, \textit{Stack Cube}, and \textit{Close Drawer}, whose linguistic descriptions are sufficiently distinctive.
Conversely, \textit{Grab Handle} and \textit{Place in Drawer} exhibit lower scores because their instructions often describe semantically similar tasks.

The confusion matrix, reported in Fig.~\ref{fig:confusion_language}, further highlights this observation.
Since the classification is based on embeddings comparison, most misclassifications occur between semantically related skills, rather than between unrelated manipulation primitives.
In particular, \textit{Grab Handle} is confused with \textit{Open Drawer} in $13\%$ of the cases, while \textit{Open Drawer} is misclassified as \textit{Place in Drawer} in $14\%$ of the samples.
Similarly, \textit{Close Drawer} is predicted as \textit{Open Drawer} in $13\%$ of the queries.
In contrast, primitives associated with different manipulation objectives, such as \textit{Pick Cube} and \textit{Stack Cube}, exhibit lower confusion with the drawer-related skills, confirming that the learned embedding space preserves meaningful semantic relationships among manipulation primitives.

\begin{table}[ht]
    \centering
        \begin{tabular}{l|c|c|c}
            \toprule
            \textbf{Skill} &
            \multicolumn{1}{c|}{\begin{tabular}{@{}c@{}}\textbf{Language}\\ \textbf{Recognition}\end{tabular}} &
            \multicolumn{1}{c|}{\begin{tabular}{@{}c@{}}\textbf{Imitation}\\ \textbf{Learning}\end{tabular}} &
            \multicolumn{1}{c}{\begin{tabular}{@{}c@{}}\textbf{Overall}\\ \textbf{Success}\end{tabular}}\\
            \midrule
            Pick Cube       & $89.0\%$    & $90.0\%$ & $80.1\%$ \\
            Stack Cube      & $85.0\%$    & $94.0\%$ & $79.9\%$ \\
            \midrule
            Grab Handle     & $82.0\%$    & $90.0\%$ & $73.8\%$ \\
            Open Drawer     & $85.0\%$    & $86.0\%$ & $73.1\%$ \\
            Place in Drawer & $93.0\%$    & $92.0\%$ & $85.6\%$ \\
            Close Drawer    & $83.0\%$    & $88.0\%$ & $73.0\%$ \\
            \bottomrule
        \end{tabular}%
     \caption{\textbf{Evaluation of language-guided diffusion planning}, in terms of language recognition accuracy, skill execution success rate, and request-to-goal success rate.}
    \label{tab:language_encodings}
\end{table}

Furthermore, to evaluate the complete language-guided manipulation pipeline, Table~\ref{tab:language_encodings} decomposes the end-to-end performance into three stages: (i) language recognition success rate, (ii) imitation learning success rate of the diffusion model, and (iii) overall request-to-goal success rate.
The language encoder correctly identifies the intended primitive with success rates between $82\%$ and $93\%$, while the diffusion model successfully executes the selected skill with success rates ranging from $86\%$ to $94\%$. Combining these two stages yields overall request-to-goal success rates between $73.04\%$ and $85.56\%$, showing that the final performance is primarily limited by the accumulation of errors across the sequential modules rather than by either component individually.

The highest end-to-end performance is obtained for \textit{Place in Drawer} ($85.5\%$), benefiting from both reliable language recognition ($93.0\%$) and accurate trajectory generation ($92\%$).

Overall, these results demonstrate that the proposed language encoder produces sufficiently discriminative semantic embeddings to reliably retrieve the intended manipulation primitive, while the integration with the diffusion model enables robust end-to-end execution of language-guided manipulation tasks.

\subsection{Language uncertainty quantification}
\label{subsec:lang_cp}
%
As introduced in Sec.~\ref{subsubsec:conf_pred}, estimating the uncertainty associated with language instruction classification is essential for reliable and safe interaction.
To mitigate potential failures arising from incorrect skill assignments, we employ Conformal Prediction (CP) to distinguish in-scope utterances corresponding to the available manipulation skills from out-of-scope requests.
In-scope utterances are classified according to the similarity of their language embeddings to the corresponding skill representations.
Conversely, requests that cannot be reliably associated with a known skill are identified as out-of-scope and forwarded to the LLM for further interpretation and task decomposition.

For the evaluation, a dataset of $900$ natural language utterances is considered, comprising $750$ single-skill requests, $50$ long-horizon task requests, and $100$ out-of-scope requests.
The dataset is divided into disjoint calibration and test sets.
The calibration set contains $150$ utterances exclusively corresponding to the available manipulation skills and is used to calibrate the CP thresholds with a significance level of $\alpha=0.1$.
The remaining $750$ utterances form the test set and are used to evaluate the ability of the proposed approach to distinguish between single-skill requests, long-horizon task requests, and out-of-distribution inputs.

The results of the CP-based classification mechanism are reported in Tab.~\ref{tab:results_llm_combined}.
All out-of-distribution (OOD) utterances are assigned an empty prediction set, indicating that none of the predefined manipulation skills can be reliably associated with these requests.
Single-skill utterances yield empty, singleton, or larger prediction sets depending on their linguistic formulation.
Only singleton prediction sets are directly classified by the embedding-based module, whereas requests resulting in empty or ambiguous prediction sets are instead forwarded to the LLM for further interpretation.

Long-horizon task requests are predominantly assigned empty prediction sets, allowing them to be correctly identified as incompatible with a single known skill and subsequently forwarded to the LLM for skill-sequence generation.
However, $20\%$ of task requests produce a singleton prediction set, causing them to be incorrectly classified as confident in-scope inputs despite requiring multiple manipulation primitives.

\begin{table}[ht]
    \centering
    \resizebox{\columnwidth}{!}{%
    \begin{tabular}{l|c|cc|cc|cc}
        \toprule
        \multirow{2}{*}{\textbf{Scenario}} & \multirow{2}{*}{\makecell{\textbf{Total}\\\textbf{Utterances}}} 
        & \multicolumn{2}{c|}{\makecell{\textbf{empty set (ood)}}}
        & \multicolumn{2}{c|}{\makecell{\textbf{singleton (certain)}}}
        & \multicolumn{2}{c}{\textbf{large set (ambiguous)}} \\
        & & \textbf{\#} & \textbf{\%} & \textbf{\#} & \textbf{\%} & \textbf{\#} & \textbf{\%} \\
        \midrule
        Skill & 600 & 175 & 29.17\% & 220  & 36.67\% & 205 & 34.17\% \\
        Task  & 50  & 40  & 80.00\% & 10   & 20.0\%  & 0  & 0.00\% \\
        OOD   & 100 & 100 & 100\%& 0     & 0.00\%  & 0 & 0.00\%\\
        \bottomrule
    \end{tabular}%
    }
    \caption{Conformal prediction: uncertainty types}
    \label{tab:results_llm_combined}
\end{table}

\begin{figure}[ht]
    \begin{center}
    \resizebox{\textwidth}{!}{\input{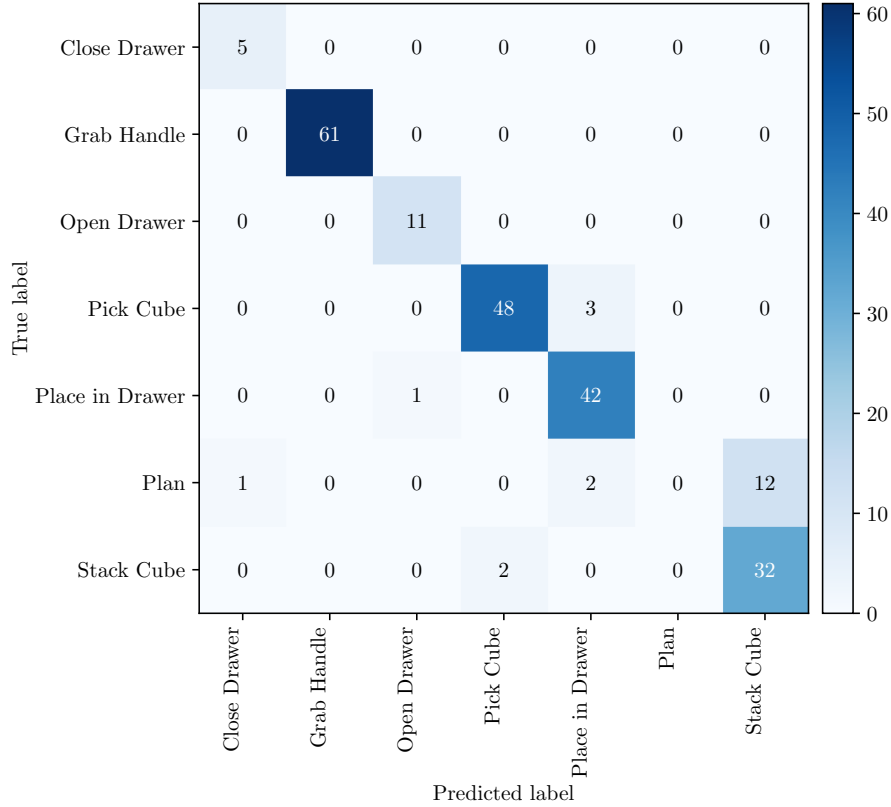}}
    \end{center}
    \caption{Confusion matrix of the language encoder on certain scenarios.}
    \label{fig:confusion_language_certain_scenarios}
\end{figure}
Whenever CP identifies an utterance as in-scope, the corresponding skill is determined based on the embedding similarity with the available manipulation primitives.
CP-certain skill predictions achieve an accuracy of $90.45\%$ ($199/220$); however, ten test instances are incorrectly classified as certain single-skill requests, despite corresponding to multi-step tasks (see the confusion matrix in Fig.~\ref{fig:confusion_language_certain_scenarios}).
As a result, the accuracy across all certain scenarios decreases to $86.52\%$, since the embedding-based model (EM) is not designed to handle multi-step task requests.




\begin{table}[ht]
    \centering
        \begin{tabular}{l|c|c|c}
            \toprule
            \textbf{Scenario} &
            \multicolumn{1}{c|}{\begin{tabular}{@{}c@{}}\textbf{Total}  \textbf{Utterances}\end{tabular}} &
            \multicolumn{1}{c|}{\begin{tabular}{@{}c@{}}\textbf{Matches}\end{tabular}} &
            \multicolumn{1}{c}{\begin{tabular}{@{}c@{}}\textbf{Success} \textbf{Rate}\end{tabular}}\\
            \midrule
            Skill   & $380$   & $274$ & $72.11\%$ \\
            Task    & $40$    & $29$  & $72.5\%$ \\
            OOD     & $100$   & $86$  & $86.0\%$ \\
            \midrule
            \textbf{Total} & $520$ & $389$ & $74.81\%$\\
            \bottomrule
        \end{tabular}%
     \caption{Performance in uncertainty scenarios are delegated to an LLM generation.}
    \label{tab:uncertain_predictions}
\end{table}
Moreover, predictions associated with uncertain scenarios are delegated to the LLM.
As reported in Tab.~\ref{tab:uncertain_predictions}, the LLM correctly classifies $74.81\%$ of uncertain utterances, with $389$ out of $520$ predictions matching the ground truth.

Considering the complete test set of $750$ utterances, the system produces $588$ correct predictions, comprising $199$ correct predictions from the embedding model and $389$ from the LLM fallback.
This corresponds to an overall accuracy of $78.40\%$.

Regarding inference time, the uncertainty quantification mechanism also reduces the computational cost by selectively routing inputs to the appropriate model.
CP assigns $31\%$ ($230/750$) of the test utterances to the embedding model and $69\%$ ($520/750$) to the LLM.
As anticipated in Sec.~\ref{subsec:impl_details}, given average inference time of $13.74$~ms and $3996$~ms for the embedding model and LLM, respectively, the estimated average latency over the complete test set is: 
$0.31 \cdot 13.74 + 0.69 \cdot 3996 = 2761.49~\text{ms}$
per interaction.
In comparison, an LLM-only baseline requires an average of $3996$~ms per interaction.
Therefore, the proposed CP-based uncertainty quantification reduces the average inference latency by approximately $1234.51$~ms per interaction.

\section{Conclusions, Limitations, and Future Work}
\label{sec:conclusion}

This paper presented DROM, a framework for learning and composing robotic manipulation skills from a limited number of human demonstrations.
By combining Dynamic Movement Primitive (DMP)-based trajectory augmentation, language-guided skill representation, and diffusion-based trajectory generation, the proposed approach enables a single generative model to execute a diverse library of manipulation primitives and compose them into long-horizon tasks directly from natural language instructions.

The experimental evaluation addressed the three research questions introduced in this work.
First, the results demonstrated that DMP-based augmentation allows a small set of expert demonstrations to be expanded into expressive training datasets, enabling robust policy learning while substantially reducing the amount of manual data collection.
Second, we showed that a single language-conditioned diffusion model can successfully represent multiple manipulation skills, including orientation-sensitive behaviors that are difficult to design with conventional motion planning or hard-coded controllers, while generalizing beyond the demonstrated workspace.
Finally, the proposed language-guided pipeline effectively translated natural language instructions into executable sequences of manipulation primitives, allowing the robot to autonomously accomplish long-horizon tasks through the composition of previously learned skills.

Overall, the proposed framework provides a unified approach for scalable robot programming, where new manipulation capabilities can be incorporated by collecting only a small number of demonstrations without retraining task-specific planners or manually designing motion strategies.
These characteristics make DROM particularly suitable for flexible industrial and collaborative robotic applications, where manipulation requirements continuously evolve and rapid deployment is essential.

Despite these results, the current framework has two main limitations.
First, the diffusion model does not explicitly account for changes in the positions of the objects involved in the manipulation task during execution.
Once a trajectory has been generated, changes in the object configuration cannot be directly incorporated into the diffusion model state to condition the generation of a new trajectory.
Consequently, the current formulation does not provide a mechanism for dynamically regenerating the trajectory based on changes in the environment or object configuration during execution.
Second, the vision and diffusion components are currently implemented as separate modules.
The vision pipeline is used to estimate the relevant scene information, while the diffusion model independently generates the manipulation trajectory.
This decoupled architecture limits the ability of the policy to continuously integrate visual observations during trajectory generation and prevents the model from directly adapting its predictions to visual changes occurring throughout the execution of a task.

Future work will focus on addressing these limitations and extending DROM toward more reactive and tightly integrated manipulation policies.
First, we plan to investigate hierarchical diffusion policies capable of jointly reasoning over skill sequencing and low-level trajectory generation within a single generative architecture.
Second, we aim to incorporate multimodal sensory feedback, including force, tactile, and visual observations, to improve robustness during contact-rich manipulation.
Third, we intend to evaluate the framework on substantially larger manipulation libraries and more complex industrial assembly scenarios involving dynamic environments, human-robot collaboration, and online adaptation to previously unseen tasks.
Finally, we will investigate the integration of learning-based object pose estimation methods, replacing fiducial markers with vision models capable of localizing previously unseen objects in unstructured environments.






\bmhead{Acknowledgements}

This research was partially supported by Horizon Europe project \emph{Fluently}, Grant agreement ID: 101058680










\begin{appendices}

\section{Dynamic Movement Primitives}
\label{app:DMPs}

In this section, we briefly review the formulation of Dynamic Movement Primitives (DMPs) as introduced in~\citep{dmps01}.

The DMP framework models a trajectory as a second-order dynamical system of the form~\citep{dmps02,dmps03}:
\begin{equation}
\label{eq:dmp_equation}
\ddot{y}(t) = \alpha_y \big(\beta_y (g - y(t)) - \dot{y}(t)\big) + f(x),
\end{equation}
where $y(t)$ denotes the system state, $\dot{y}(t)$ and $\ddot{y}(t)$ its first and second derivatives, $\alpha_y, \beta_y > 0$ are gain parameters, and $g$ is the goal state. The term $f(x)$ is a nonlinear forcing function that encodes the demonstrated motion.

The forcing term is defined as a normalized weighted sum of $N$ radial basis functions:
\begin{equation}
\label{eq:forcing_term}
f(x) = \frac{\sum_{i=1}^{N} \Psi_i(x)\,\omega_i}{\sum_{i=1}^{N} \Psi_i(x)}\, x (g - y_0),
\end{equation}
where $y_0$ is the initial state, $\omega_i$ are learnable weights, and $\Psi_i(x)$ are Gaussian basis functions:
\begin{equation}
\label{eq:psi}
\Psi_i(x) = \exp\big(-h_i (x - c_i)^2\big).
\end{equation}

In all experiments, we employ Gaussian radial basis functions with $N=200$ basis functions uniformly distributed along the phase variable. The spring--damper gains are set to $\alpha_y=25$ and $\beta_y=\alpha_y/4=6.25$, yielding a critically damped system that ensures smooth convergence to the goal while preserving the demonstrated motion.

The phase variable $x$ is governed by a canonical system:
\begin{equation}
\label{eq:canonical_system}
\dot{x} = \alpha_x x,
\end{equation}
which ensures a monotonic decay over time.

The parameters $\omega_i$ are learned from demonstrated trajectories. Given a demonstration
$\tau_{\text{demo}} = \{y_d(t_0), y_d(t_1), \dots, y_d(t_f)\}$, the target forcing term is obtained by inverting~\eqref{eq:dmp_equation}:
\begin{equation}
\label{eq:f_d}
f_d(t) = \ddot{y}_d(t) - \alpha_y \big(\beta_y (g - y_d(t)) - \dot{y}_d(t)\big).
\end{equation}

Learning is then formulated as a weighted regression problem. For each basis function $\Psi_i$, Locally Weighted Regression (LWR) estimates $\omega_i$ by minimizing:
\begin{equation}
\label{eq:Ji}
J_i = \sum_{t=1}^{P} \Psi_i(t)\,\big(f_d(t) - \omega_i\, x(t)(g - y_0)\big)^2.
\end{equation}

Once learned, the weights $\{\omega_i\}_{i=1}^{N}$ allow the reproduction of the demonstrated trajectory. Moreover, the DMP formulation enables generalization to novel initial and goal states while preserving the structure of the demonstrated motion, thereby ensuring adaptability across different task configurations.

\section{Prompt structure}\label{prompt}
Follows the structure of the LLM prompt. 
The prompt comprises a general presentation of the primitives, the list of the available objects, a specification of the parameters each primitive involves and a set of examples with input and expected output.
\begin{lstlisting}
You are Franka Emika robotic arm. 
Your task is to identify the task plan that a sequence of actions to respond to a certain query of the human collaborating with you.

This is the list of you possible actions:


"grab_handle"       grasp the handle of the drawer, it is necessary before open it
"open_drawer"       open the drawer
"close_drawer"      close the drawe
"pick"              pick any cube on the table
"place_in_drawer"   place the object taken in the drawer  
"place"             place something in a specific place 

                    
This is the list of available objects on the scene:


"table"
"red"
"blue"
"green"
"black"
"purple"
"gray"
"yellow"
"cup"
"bottle"
"battery"


notice that some actions have specific parameter that take in input:


"grab_handle"       no parameter
"open_drawer"       no parameter
"close_drawer"      no parameter
"pick"              pick(<color>) has one parameter; it refers to the color of the cube to grasp or to an object
"place_in_drawer"   no parameter
"place"             place(<location>) has one parameter; it refers to the location where to place the object. Assume you already have an object in the end-effector and you want to place it either on the table or on another colored cube.
"pour"              pour(<location>) has one parameter; it refers to the location where pour a liquid. 

these are a few examples of what you have to do:       


Query: Stack the green cube on the yellow
{
    "thinking": "the green cube should be on top, so I pick with a pick(green) and place it on the yellow place(yellow) to form a stack",
    "plan": [
        {"action": "pick", "object":"green"},
        {"action": "place", "object":"yellow"}
    ]
}

Query: Take the black cube and place it on the table
{
    "thinking": "to take the black cube I must pick it with pick(black) and to to put it on the table I must place(table)",
    "plan": [
        {"action": "pick", "object":"black"},
        {"action": "place", "object":"table"}
    ]
}

Query: Open the drawer
{
    "thinking": "to open the locker i should take its handle with grab_handle and the open it with open_drawer",
    "plan": [
        {"action": "grab_handle"},
        {"action": "open_drawer"}
    ]
}

Query: Put the red cube in the drawer
{
    "thinking": "to put the red cube into the drawer i should open it first grabbing the handle and then open it. Then I can pick the red cube and place it in the locker, optionally I can close it",
    "plan": [
        {"action": "grab_handle"},
        {"action": "open_drawer"},
        {"action": "pick", "object":"red"},
        {"action": "place_in_drawer"},
        {"action": "close_drawer"}
    ]
}

Query: Close the drawer
{
    "thinking": "simple! I just to the action close_drawer",
    "plan": [
        {"action": "close_drawer"}
    ]
}

Input that do not refer to any of the available skills or composition of the available skills do not need to be addressed.
Do not provide an output in this case.

be sure to output only a valid json and nothing else, not even ```json or something similar.
return as JSON.
\end{lstlisting}



\end{appendices}


\bibliography{biblio}

\end{document}